\pdfoutput=1

\documentclass[11pt]{article}

\usepackage[final]{acl}

\usepackage{times}
\usepackage{latexsym}

\usepackage[T1]{fontenc}

\usepackage[utf8]{inputenc}

\usepackage{microtype}

\usepackage{inconsolata}

\usepackage{graphicx}

\title{Enhancing Rating Prediction with Off-the-Shelf LLMs\\Using In-Context User Reviews}

\author{
  Koki Ryu${}^{1, 2}$ \quad
  Hitomi Yanaka${}^{1, 2}$ \\
  ${}^1$The University of Tokyo \\
  ${}^2$Riken \\
  \texttt{\{kokiryu, hyanaka\}@is.s.u-tokyo.ac.jp}}

\usepackage{comment}
\usepackage{amsmath}
\usepackage{cuted}
\usepackage{titlesec}
\usepackage{booktabs}    %
\usepackage{tabularx}    %
\usepackage{subcaption}
\usepackage{xcolor}
\usepackage[most]{tcolorbox}
\usepackage{hyperref}

\newcommand{\rstors}{RS $\to$ RS }
\newcommand{\rstorsnospace}{RS $\to$ RS}
\newcommand{\rstos}{RS $\to$ S }
\newcommand{\rstosnospace}{RS $\to$ S}
\newcommand{\stos}{S $\to$ S }
\newcommand{\stosnospace}{S $\to$ S}
\newcommand{\tos}{$\emptyset \to$ S }
\newcommand{\tosnospace}{$\emptyset \to$ S}
\newcommand{\highlight}{\color{black} } %
\newcommand{\finishhighlight}{\color{black}}

\newenvironment{Prompt}[1][Prompt Example]{
  \begin{tcolorbox}[
    colback=white,
    colframe=black!50,
    boxrule=0.5pt,
    arc=2pt,
    outer arc=2pt,
    width=\linewidth,
    enlarge top by=5pt,
    enlarge bottom by=5pt
  ]
  \small
}{
  \end{tcolorbox}
}

\titlespacing\paragraph
  {0pt}
  {0.8ex plus 0.5ex minus .2ex}
  {0.8em}

\begin{document}
\maketitle
\begin{abstract}
Personalizing the outputs of large language models (LLMs) to align with individual user preferences is an active research area.
However, previous studies have mainly focused on classification or ranking tasks and have not considered Likert-scale rating prediction, a regression task that requires both language and mathematical reasoning to be solved effectively.
This task has significant industrial applications, but the utilization of LLMs remains underexplored, particularly regarding the capabilities of off-the-shelf LLMs.
This study investigates the performance of off-the-shelf LLMs on rating prediction, providing different in-context information.
Through comprehensive experiments with eight models across three datasets, we demonstrate that user-written reviews significantly improve the rating prediction performance of LLMs.
This result is comparable to traditional methods like matrix factorization, highlighting the potential of LLMs as a promising solution for the cold-start problem.
We also find that the reviews for concrete items are more effective than general preference descriptions that are not based on any specific item.
Furthermore, we discover that prompting LLMs to first generate a hypothetical review enhances the rating prediction performance.
Our code is available at \url{https://github.com/ynklab/rating-prediction-with-reviews}.
\end{abstract}

\section{Introduction}
Recent large language models (LLMs) have demonstrated remarkable capabilities across various tasks without task-specific fine-tuning, one of which is personalization.
By providing in-context user preference data and applying prompt engineering techniques, previous studies enabled off-the-shelf LLMs to align with individual preferences in tasks such as preferred item prediction \cite{gpg}, top-N recommendation \cite{top-n-evaluation}, and item reranking \cite{recommender-comprehensive, zero-shot-rankers}. \par

However, these studies have not fully covered all personalization tasks.
An important example is \textbf{Likert-scale rating prediction}.
As represented by notable examples such as Netflix Prize \cite{netflix} and MovieLens \cite{movielens}, rating prediction has been one of the key areas of recommendation, though the application of off-the-shelf LLMs to it remains underexplored.\par
\begin{figure}[t]
  \centering
    \includegraphics[width=1.0\columnwidth]{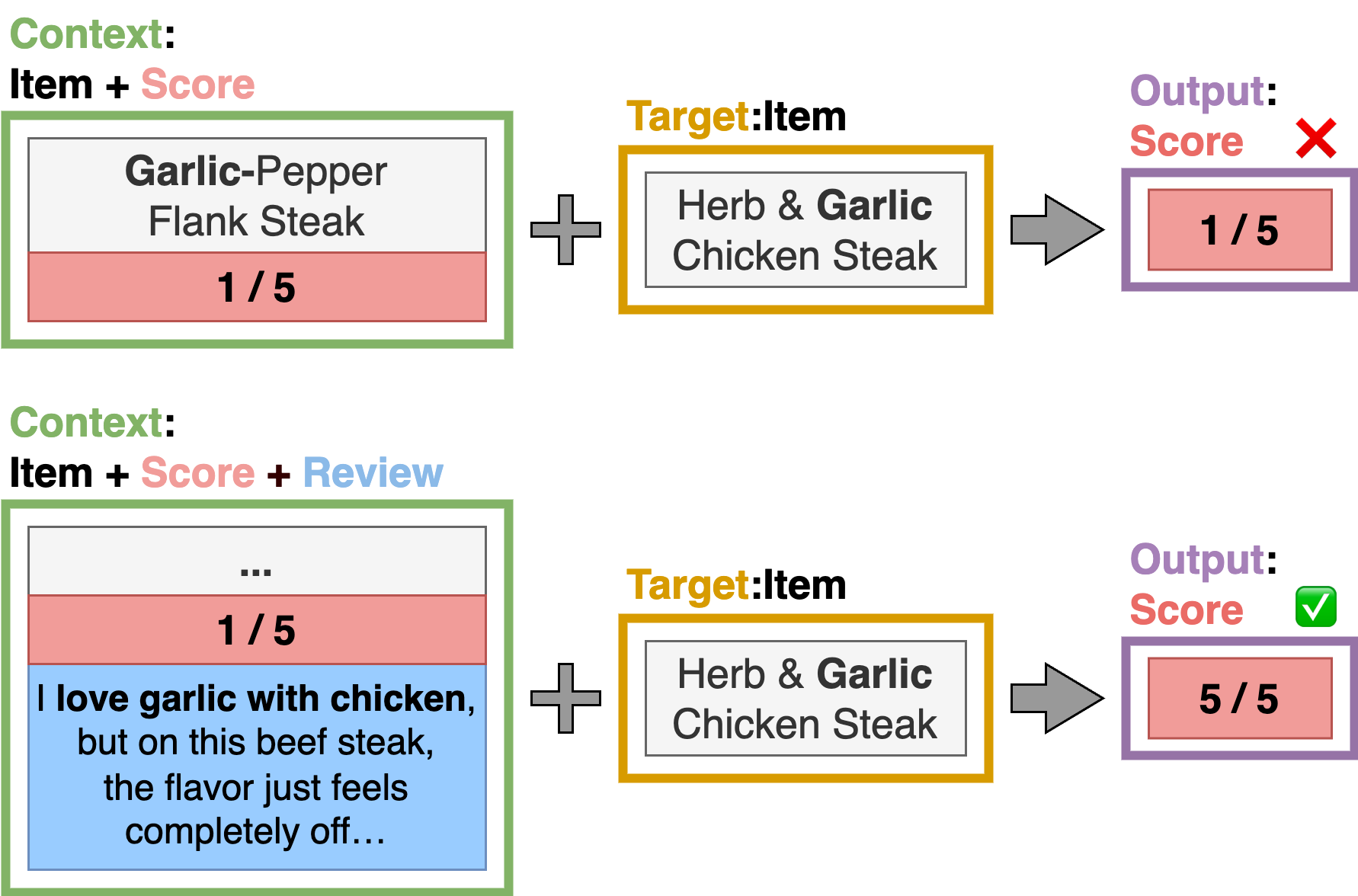}
  \caption
  {Illustration of the impact of in-context review data on LLM-based rating prediction performance.
  By leveraging rich, qualitative preference information from user reviews in the context, LLMs can more accurately infer a user’s preference for the target item, as demonstrated by the improved prediction from 1/5 to 5/5.
  }
  \label{fig:base-diagram}
\end{figure}

Rating prediction presents unique challenges, as models need to consider multiple factors such as the trend of the users' ratings or the average ratings for the items.
Therefore, traditional methods such as matrix factorization \cite{mf} require
a large volume of interaction histories for both a user and an item to be effective, which can cause the so-called  ``cold-start problem''~\cite{cold-start-problem, llm-personalization-survey}, where performance is degraded for new users or items with insufficient data.\par

To address this, some studies have applied fine-tuned LLMs to the rating prediction tasks~\cite{llm-user-rating-pred, permpst}.
In particular, \citet{permpst} demonstrated that using user-written review texts as in-context information allowed a fine-tuned model to achieve reasonable performance with just a few inference-time examples. 
This is because review texts contain rich qualitative details about the user's preferences that cannot be captured by numerical scores alone, as illustrated in \autoref{fig:base-diagram}. \par
Extending this review-based approach to off-the-shelf LLMs would enable the development of a more lightweight rating prediction system that does not require costly domain-specific fine-tuning data.
Furthermore, observing the behavior of these general-purpose models on this complex task would enhance our understanding of their personalization capabilities.\par

In this paper, we investigate how different forms of in-context information contribute to the rating prediction performance by off-the-shelf LLMs.
First, we benchmark the rating prediction performance of eight open and closed LLMs with and without the in-context user-written reviews. 
Second, we compare the effectiveness of per-item reviews against another format of preference data utilized in prior work~\cite{profile-based-recommendation}.
Finally, we explore the prompting strategies to further exploit the information in the provided reviews.

Our key findings are the following:
\begin{itemize}
    \setlength{\parskip}{0cm}
    \setlength{\itemsep}{0cm}
    \item Per-item review texts consistently improve the user rating prediction performance across diverse datasets and LLMs, including open and closed models.
    In particular, OpenAI o3\footnote{https://openai.com/index/introducing-o3-and-o4-mini/} achieves an absolute improvement of 0.147 in Spearman correlation and 13.0\% relative reduction in Root Mean Squared Error (RMSE) on the Per-MPST~\cite{permpst} dataset. 
    \item Per-item review texts enhance the rating prediction of LLMs better than the general preference description not grounded in specific items, which is the in-context information used by  \citet{profile-based-recommendation}.
    \item 
    Instructing an LLM to first generate a hypothetical review before predicting a score is a promising strategy to enhance performance further.
    \highlight
    This effect is particularly pronounced on smaller models.
    \finishhighlight
\end{itemize}

\section{Related Work}

\subsection{Personalization with Off-the-Shelf LLMs}
Many previous studies have explored using off-the-shelf LLMs for personalization, primarily by leveraging users' historical interactions.
For instance, \citet{zero-shot-rankers} used off-the-shelf LLMs for the personalized item ranking task.
Similarly, \citet{top-n-evaluation} evaluated ChatGPT\footnote{https://openai.com/index/chatgpt/}'s performance on the top-N recommendation task based on historical interaction. 
\citet{user-profile-experiment} demonstrated that providing a user's historical responses in the context improves an off-the-shelf LLM's performance on the LaMP~\cite{lamp} dataset. 
\citet{gpg} proposed a technique where an LLM is instructed to summarize a user's historical responses in a specific manner to improve the performance of off-the-shelf LLMs on the multiple-choice preference prediction task.
\citet{recommendation-potential-analysis} conducted a large-scale performance analysis across different LLMs on item reranking tasks based on historical interactions.\par
Another stream of research focuses on preferences explicitly described in text.
\citet{narrative-recommendation} proposed a recommendation system based on free-form text user requests using off-the-shelf LLMs and basic prompt engineering techniques, such as few-shot or role-playing prompting.
\citet{profile-based-recommendation} collected self-described preferences of users to enhance the item reranking performance by LLMs. 
However, these studies are limited to simpler tasks such as top-N recommendation or item reranking.
Thus, whether these approaches can be directly applied to the more difficult rating prediction task remains unclear.
Furthermore, the specific effect of per-item review texts has not been thoroughly investigated.

\subsection{Rating Prediction with LLMs}
While the use of off-the-shelf LLMs on the rating prediction task is not well-investigated, several studies have utilized fine-tuned models for this task.
For example, \citet{llm-user-rating-pred} reported that fine-tuned LLMs could achieve rating prediction performance comparable to traditional recommender systems.
\citet{permpst} proposed Per-MPST, a rating prediction dataset with past review texts available as in-context input.
They also proposed PerSE as the framework for solving the problem with fine-tuned LLMs and achieved reasonable prediction performance with a few in-context examples.
This finding leads us to investigate whether off-the-shelf LLMs can reproduce the positive effect with review texts, which could eliminate the need for costly fine-tuning.

\subsection{Prompt Engineering on Personalization}
\label{subsection:prompt-engineering-related-work}
Prompt engineering is a crucial technique to enhance LLM performance on various domains.
Chain-of-thought (CoT;~\citealp{cot, zero-shot-cot}) is one of the most notable examples. 
However, its benefit may be limited, as a recent study~\cite{not-cot} suggests that CoT only works effectively on domains that require mathematical or logical reasoning.\par
Given the limitations of such general-purpose methods, task-specific prompting strategies have been proposed in the personalization domain.
\citet{gpg} instructs LLMs to generate intermediate outputs from a specific viewpoint. 
The prompt used by \citet{permpst} has the LLMs explicitly write down a hypothetical review from the user's perspective. 
Another line of work, such as
Knowledge Augmented Generation (KAR;~\citealp{kar}), LLM-Rec~\cite{llm-rec}, and UR4Rec~\cite{ur4rec} generate intermediate texts with LLMs to increase the input data to the fine-tuned recommendation models.
However, the effectiveness of these prompting techniques remains untested specifically for the rating prediction tasks with off-the-shelf LLMs.

\section{Problem Formulation}
In this section, we formally define the task of rating prediction using off-the-shelf LLMs.
We specifically focus on evaluating the effect of providing user-written reviews as in-context data.
The task is formulated as follows.\par
Let the target LLM be $\mathcal{M}$.
Given a user $u$ and a target item with description $x_u$, the goal is to have the model $\mathcal{M}$ predict the numerical rating $y_u$ that $u$ would assign to the item represented by $x_u$. 
The ground-truth rating $y_u$ is an integer within the range $[y_{min}, y_{max}]$, where $y_{min}$ and $y_{max}$  are dataset-specific parameters that denote the minimum and maximum scores.\par
For each prediction, the model $\mathcal{M}$ is provided with two additional inputs: $p_u$, a set of texts that contains $u$'s personal preference information, such as $u$'s past review history (referred to as the user profile), and $I$, an instruction that specifies the input and output formats of the task. 
Based on those inputs, $\mathcal{M}$ generates a raw text output $o_u$ as:
\begin{align}
    o_u = \mathcal{M}(I, x_u, p_u)
\end{align}
Since the raw output $o_u$ could contain additional texts other than the predicted rating, we define an instruction-specific extraction function $f_I$ to parse the final predicted score $y_u'$ as:
\begin{align}
    y_u' = f_I(o_u).
\end{align}
We define our evaluation dataset as $\mathcal{D} = \{(x_u, p_u, y_u)\}_{u \in \mathcal{U}}$ for a set of users $\mathcal{U}$.
The final performance is measured by comparing the set of predicted ratings and ground-truth ratings $\{(y'_u, y_u)\}_{u \in \mathcal{U}}$.
In this work, we control the content of the user profile $p_u$ and the instruction $I$ to investigate their effects on prediction performance.

\section{General Setup}
\subsection{Datasets}
\label{base-datasets}
We evaluate the models on three distinct datasets to assess the performance across different domains. \par
Per-MPST~\cite{permpst} (Movies) is a movie review dataset derived from the IMDb\footnote{https://www.imdb.com/} data.
Each data point consists of the movie plot, a user-written review, and a corresponding rating on a scale of 1 (lowest) to 10 (highest). 
The dataset provides multiple data splits based on the number of in-context examples ($k$); we use $k = 5$ test split in our experiments.\par
We also adapt two datasets from other domains, Recipe~\cite{recipe} and the Book category of Amazon Reviews'23~\cite{amazon-books} (Books), to match the format of the Movies dataset.
We construct a unified ``item description'' for each dataset by concatenating relevant features: (name, description, steps) or Recipe, and (title, subtitle, and features) for Books.
We then filter for reviews with at least 200 characters and randomly sample 1,000 instances for each dataset.
Each instance consists of five historical reviews and one target review for prediction, imitating the structure of the Movies dataset.
Appendix \ref{appendix:dataset-stats} provides detailed statistics for each dataset and a verification of our sampling method.

\subsection{Models}
\label{models}
To ensure comprehensive evaluation, our experiments leverage eight models, including five open-source and three closed-source LLMs. \par
For open-source models, we choose instruction-tuned versions of models widely used in recommendation research, covering various parameter sizes: Llama 3.1 8B, Llama 3.3 70B~\cite{llama-3}, Gemma 3 12B, Gemma 3 27B~\cite{gemma-3}.

We also include Qwen3 8B~\cite{qwen3} to assess the performance of models designed for reasoning tasks.
From the closed-source domain, we evaluate three state-of-the-art models: OpenAI's o3 and GPT-4.1\footnote{https://openai.com/index/gpt-4-1/}, and Anthropic's Claude Sonnet 4\footnote{https://www.anthropic.com/news/claude-4}.
Detailed model configurations are available in Appendix \ref{appendix-models}.

\subsection{Evaluation Methods}
\label{evaluation-methods}
We evaluate model performance using three metrics commonly used to measure rating prediction performance.
We use the Spearman and Kendall-Tau correlation coefficients to measure the rank correlation between predicted and ground-truth ratings.
To quantify the absolute prediction error, we use the Root Mean Squared Error (RMSE).
\par
To account for the inherent stochasticity in our model configuration, we conduct multiple runs for the open-source models.
We report the mean and standard deviation over six runs for the main experiments presented in \autoref{subsection:base-prompting-results} and \autoref{subsection:prompt-engineering-result}.
Furthermore, we conduct Welch's t-test for these experiments to measure the statistical significance of the observed performance differences between prompting methods ($p < 0.05$).
\highlight
Additionally, we analyze the prediction variance for each test instance across multiple runs.
These results are reported in Appendix \ref{appendix:variance-multiple-runs}.
\finishhighlight
\par

In some cases, model outputs could not be parsed as a valid integer score (e.g., due to generation loops). 
As the number of these instances was minimal, we excluded them from the evaluation when calculating metrics.
The detailed results, including the parse failure rate, are reported in Appendix \ref{appendix:results-detail}.

\begin{figure*}[!t]
  \centering
    \includegraphics[width=\textwidth]{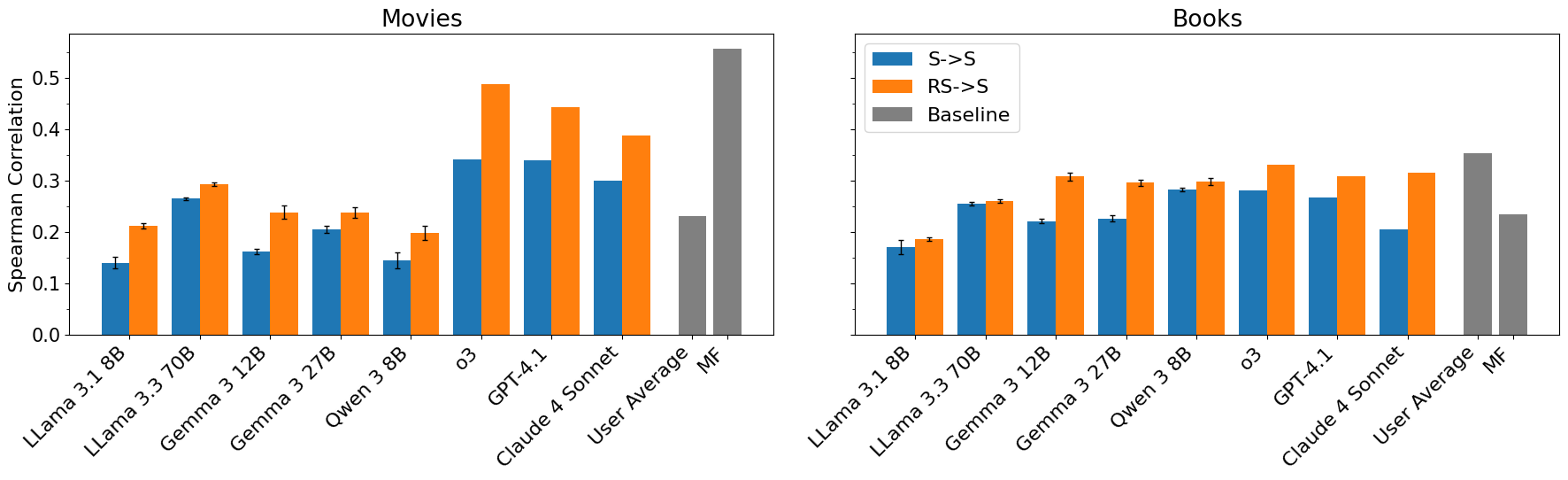}
    \includegraphics[width=\textwidth]{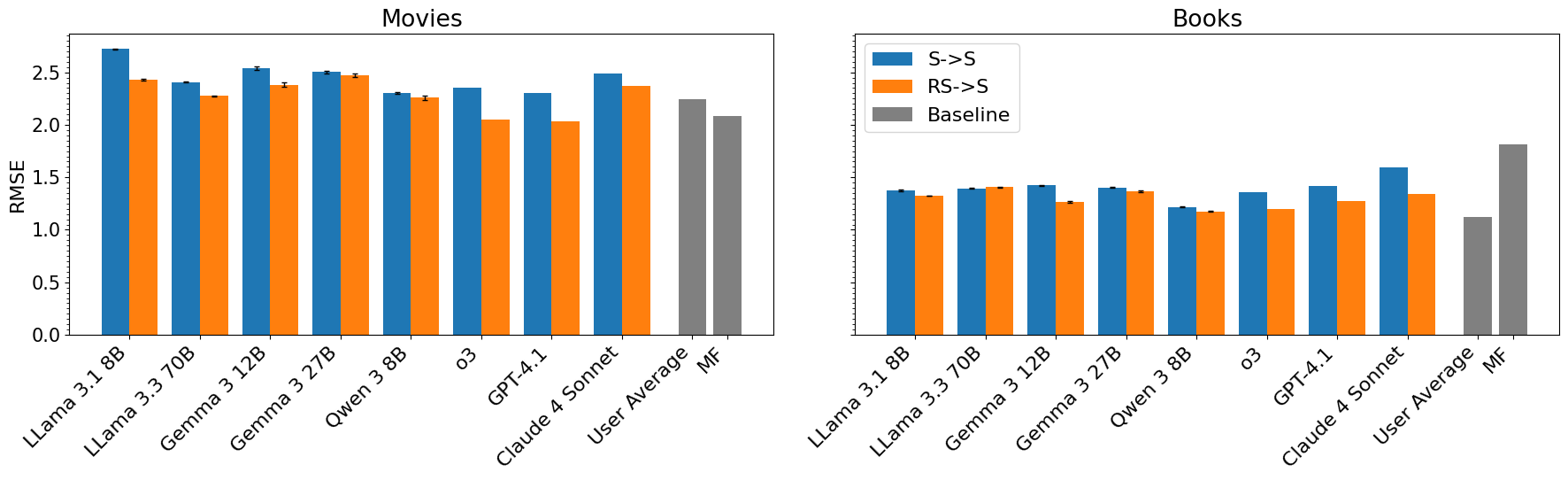}
  \caption{Average Spearman Correlation ($\uparrow$) (top) and RMSE ($\downarrow$) (bottom) with \stos and \rstos prompting.
  For the open-source models, error bars represent the standard deviation.
  \rstos format consistently improves the Spearman Correlation, while reduces the RMSE.}
  \label{fig:base-prompting-results}
\end{figure*}

\section{Effect of Review Texts}
\label{section:base-prompting}
\subsection{Comparison Method}
First, we verify whether the per-item review texts improve the personalization performance by off-the-shelf LLMs.
We accomplish this by comparing model performance under two distinct prompting formats. \par
The first format is Score-to-Score \textbf{(\stosnospace)}, where the user profile $p_u$ contains only past numerical ratings and their corresponding item descriptions.
This can be formulated as $p_u = \{(x_u^{(i)}, y_u^{(i)})\}_{i = 1}^k$, where $x_u^{(i)}$ is the description of $i$-th item and $y_u^{(i)}$ is the numerical rating user $u$ previously assigned.
The LLM is prompted to output only the predicted rating $y'_u$. \par
The second format is Review+Score-to-Score \textbf{(\rstosnospace)}, where we enrich the user profile $p_u$ with past review texts.
The profile is formulated as $p_u = \{(x_u^{(i)}, t_u^{(i)},  y_u^{(i)})\}_{i = 1}^k$, where $t_u^{(i)}$ is the user's textual review for item $x_u^{(i)}$.
In this format, the LLM also outputs only the predicted score.
Detailed prompt templates for both formats are provided in Appendix \ref{appendix:prompts}.

\subsection{Baselines}
To examine the effectiveness of the LLM-based approaches, we compare them against two traditional baseline methods following \citet{permpst}.\par
The first baseline is User Average, which predicts the target rating using the mean of the user's past ratings provided in the context.
While this method captures the user's general scoring tendency, it ignores the features of the target item.\par
The second baseline is Matrix Factorization (MF) with Alternating Least Squares (ALS)~\cite{mf}.
\highlight
MF works by representing each user and item with a low-dimensional latent vector.
These vectors are learned from the existing rating matrix such that their dot product approximates the original ratings.
\finishhighlight
Since MF requires a training phase to learn these vectors, we train it on all the in-context data available in our test set, including data from non-target users.

\subsection{Overall Results}
\label{subsection:base-prompting-results}
\autoref{fig:base-prompting-results} visualizes the performance of all models on the Movies and the Books datasets.
Full numerical results, including baseline methods, are available in Appendix \ref{appendix:results-detail}. The results show a consistent trend of performance improvement via user-written reviews provided in the \rstos prompt.
Across all 24 model-dataset combinations, the \rstos format improves the mean Spearman correlation compared to the \stos format.
A reduction in mean RMSE is also observed in 21 out of 24 combinations. \par
This improvement is also statistically significant.
Among the 15 combinations subjected to multiple runs, 14 show a statistically significant improvement in Spearman's correlation.
Similarly, 12 of the 15 combinations, except for the three models on the Recipe dataset, show a statistically significant reduction in RMSE.
These findings suggest that in-context reviews consistently enhance rating prediction performance across different models, datasets, and metrics.
The o3 model demonstrates particularly noteworthy improvements.
It achieves an absolute improvement of 0.147 in Spearman correlation and 13.0\% relative reduction in RMSE on the Movies dataset.\par
While the improvements are not statistically significant in every case, these instances are limited to the Books and Recipe datasets. Importantly, no major performance degradation was observed in any combination.
A detailed analysis of these specific cases is provided in Appendix \ref{appendix:results-detail}.

\subsection{Analysis}

\paragraph{Comparison with Baselines}
A comparison with traditional baselines further highlights the effectiveness of the user-written reviews as in-context information.
The o3 model with in-context reviews achieves smaller RMSE than both MF and user average baselines on the Movies dataset.
In particular, its superiority over MF, which was trained on significantly more ratings data, demonstrates the data-efficiency of the in-context review information.
These results underscore the potential for off-the-shelf LLMs to be a lightweight alternative to traditional methods. \par

However, an exception is observed on the Books and Recipe datasets, where LLMs with \rstos underperform the simple User Average Baseline.
As Appendix \ref{appendix:dataset-stats} shows, the rating distributions in these datasets are heavily skewed towards the maximum score.
This suggests that a simple heuristic of always predicting a high rating could be more effective for such skewed data than approaches based on the qualitative factors extracted from user reviews.
We leave a deeper analysis of this phenomenon to future work.

\paragraph{Extrapolation Capability}
\begin{figure}[tb]
  \centering
    \includegraphics[width=\columnwidth]{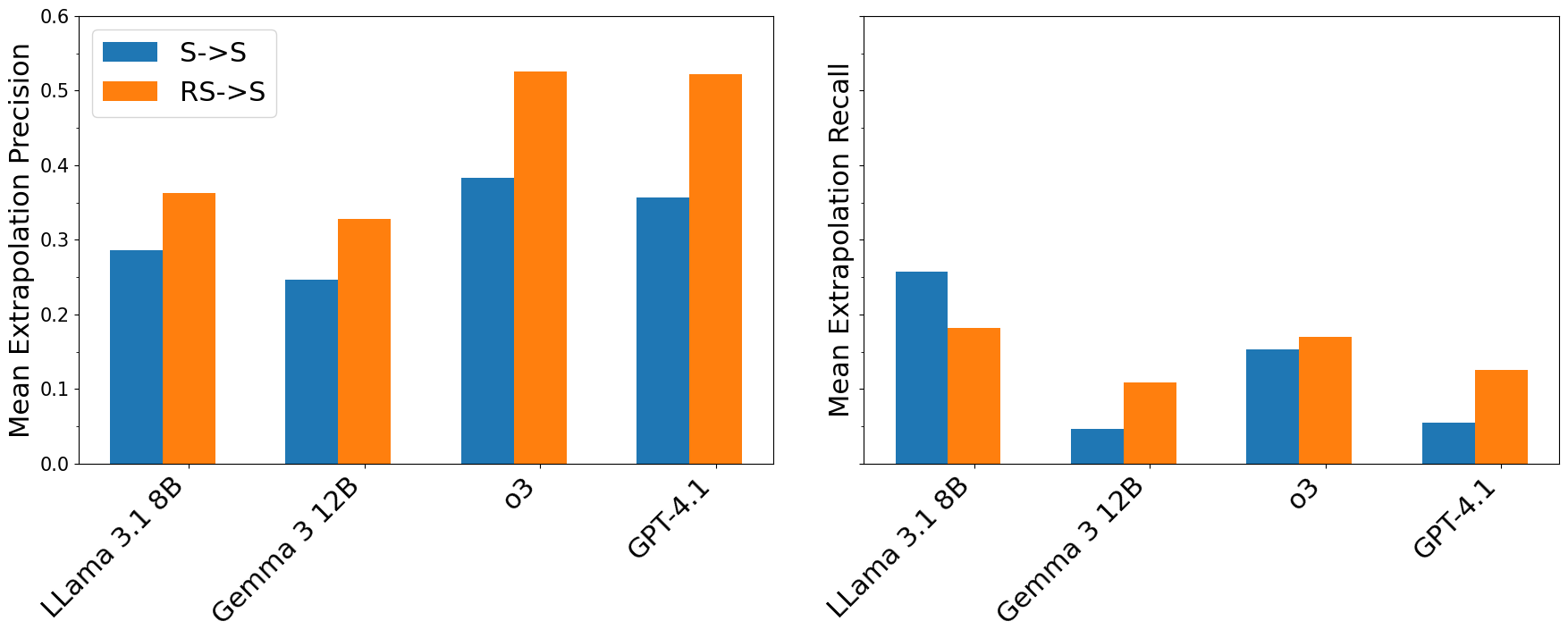}
  \caption{Comparison of extrapolation precision / recall on the Movies dataset. The models show reasonable precision, and the in-context review data improves the performance.}
  \label{fig:extrapolation-small}
\end{figure}

Unlike traditional methods such as MF and User Average, which are inherently constrained by the range of observed ratings, LLMs can theoretically predict scores outside the range of in-context examples. To verify this extrapolation capability, we analyze the predictions made in \autoref{subsection:base-prompting-results}.\par

We reframe the evaluation to assess whether models can accurately predict ratings beyond the in-context range. Performance is measured using precision and recall, specifically on instances that require such extrapolation.\par

\autoref{fig:extrapolation-small} presents selected results for the Movies dataset. As shown, while recall remains low, the models achieve reasonable precision. Furthermore, the comparison between \stos and \rstos demonstrates that in-context review texts boost this performance in most cases. This result highlights another potential advantage of LLMs over conventional methods. A more detailed analysis is provided in Appendix \ref{appendix:extrapolation}.

\paragraph{Effect of In-Context Item Similarity}
\begin{figure}[tb]
  \centering
    \includegraphics[width=\columnwidth]{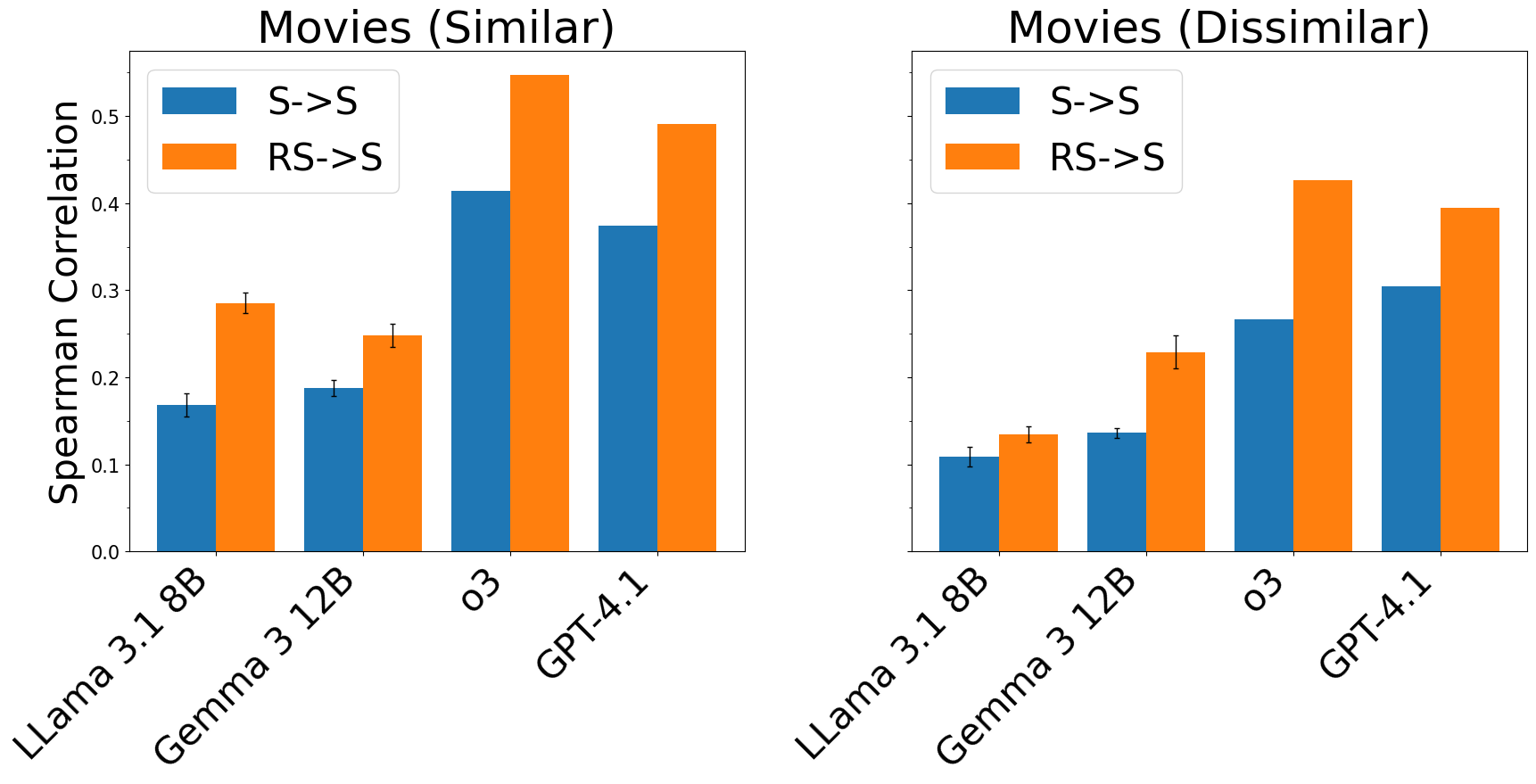}
  \caption{Comparison of Spearman Correlation on Movies (Similar) and Movies (Dissimilar). Although on the Similar subset the models perform better, in-context review texts improve the performance even on the dissimilar subset.}
  \label{fig:compare-similarity}
\end{figure}

To disentangle the effects of review texts from the similarity of in-context items, we conducted an additional analysis.
We split the Movies dataset into two parts based on the cosine similarity between the target item and the in-context items, measured using embeddings from SimCSE~\cite{princeton-roberta}\footnote{https://huggingface.co/princeton-nlp/sup-simcse-roberta-large}. This process yields ``Similar'' and ``Dissimilar'' subsets, with 351 instances each.\par
As shown in \autoref{fig:compare-similarity}, models consistently perform better on the ``Similar'' subset.
This result confirms the importance of item similarity and indicates the potential for further performance improvement via retrieval-augmented methods.\par
However, the positive effect of the review texts is observed even within the ``Dissimilar'' subset.
On this subset, adding reviews improved Spearman correlation for all eight models (with the improvement being statistically significant for all open-source models) and reduced RMSE for seven out of the eight models.
This result suggests that the in-context review texts are beneficial for LLM-based rating prediction, even when the provided in-context items are not similar to the target item.\par
More notably, for several models, including Gemma 3 12B, o3, and GPT-4.1, the performance on the ``Dissimilar'' subset with reviews surpasses the performance on the ``Similar'' subset without reviews.
This finding suggests that the textual reviews enable LLMs to infer users' preferences beyond a specific item category.

\section{Comparison with Other In-Context Preference Information}
\label{section:self-described-preference}
In this section, we compare the effect of per-item review texts with another form of in-context preference information utilized in prior work on LLM-based personalization.
Specifically, we evaluate ``self-described preferences'', a form of preference information used by \citet{profile-based-recommendation} to enhance the performance of LLM-based top-N prediction.
Unlike the per-item user reviews we focus on in this paper, self-described preference is a free-form text where users describe their general preferences without directly referencing specific items (e.g. ``I like action movies...'').
See the detailed difference between the two data types in \autoref{fig:self-description-diagram} of Appendix~\ref{appendix-self-description}.

\subsection{Settings}
\paragraph{Dataset Synthesis}
A direct comparison of the two data types is challenging, as existing datasets generated by humans do not contain both per-item reviews and self-described preferences.
To bridge the gap, we synthesize self-described preferences by prompting LLMs (Llama 3.1 8B and Gemma 3 12B) to summarize the preferences described in the available per-item reviews.
Each synthetic self-described preference is generated using the same model that performs the final prediction. 
\highlight
The generation workflow is verified in Appendix \ref{appendix-self-description-generation}, where we show that the choice of the preference-generator model does not significantly impact the conclusions.
\finishhighlight
The full implementation details and generation examples are provided in Appendix \ref{appendix-self-description}.

\paragraph{Prompting Methods}
To evaluate the effect of these synthesized preferences, we test them both in isolation and in combination with per-item examples.\par
First, to assess the self-described preference in isolation, we use a format where the user profile $p_u$ contains only the synthesized preference text.
We refer to this as the Description + None-to-Score (\tosnospace) setting, which tests if the synthesized preference summary is sufficient for prediction.\par
Second, we investigate its effect as supplementary information. We combine the synthesized preference description with our main prompting formats (\stosnospace, \rstosnospace) by appending the self-described preference text to user profile $p_u$.
This tests whether the two types of preference data are complementary.

\subsection{Results and Analysis}
\label{self-description}
\begin{figure}[t]
  \centering
    \includegraphics[width=1.0\columnwidth]{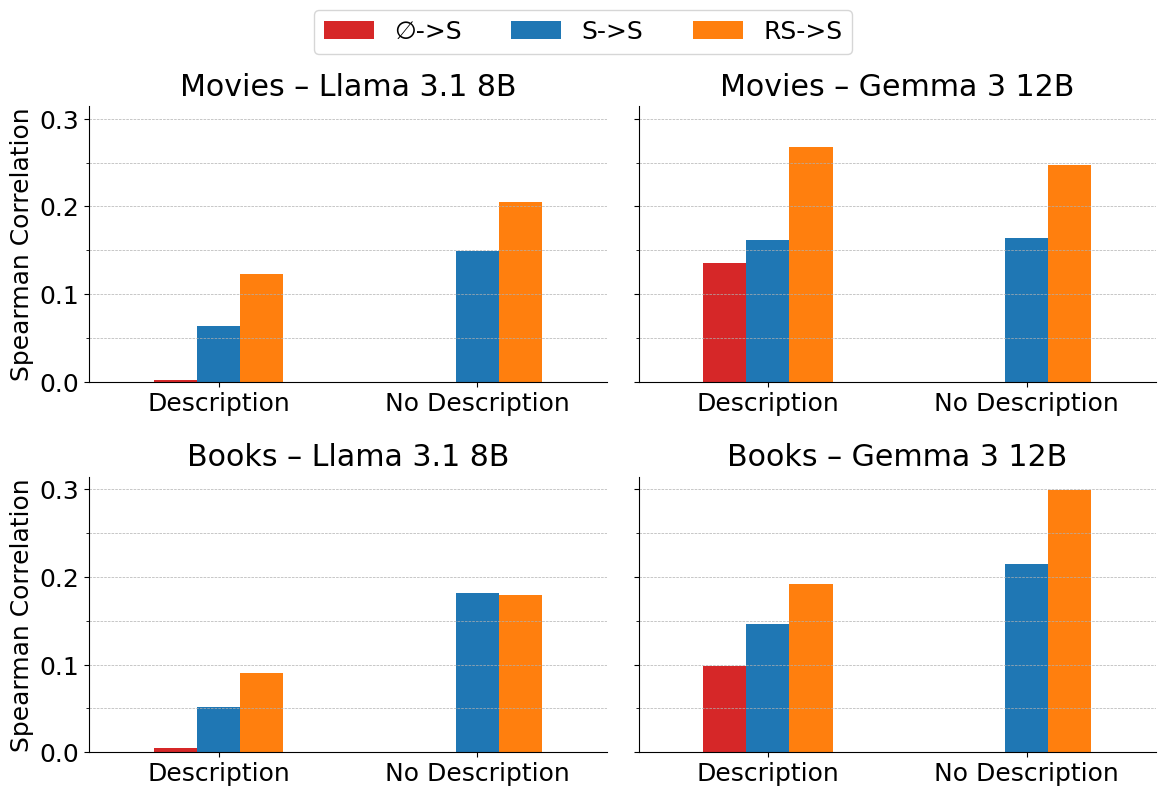}
  \caption{Comparison of rating prediction performance with and without the self-described preference generated by LLMs. The self-described preference does not work as effectively as the per-item reviews under the rating prediction settings.}
  \label{fig:self-description-results}
\end{figure}

\begin{figure*}[!t]
  \centering
    \includegraphics[width=\textwidth]{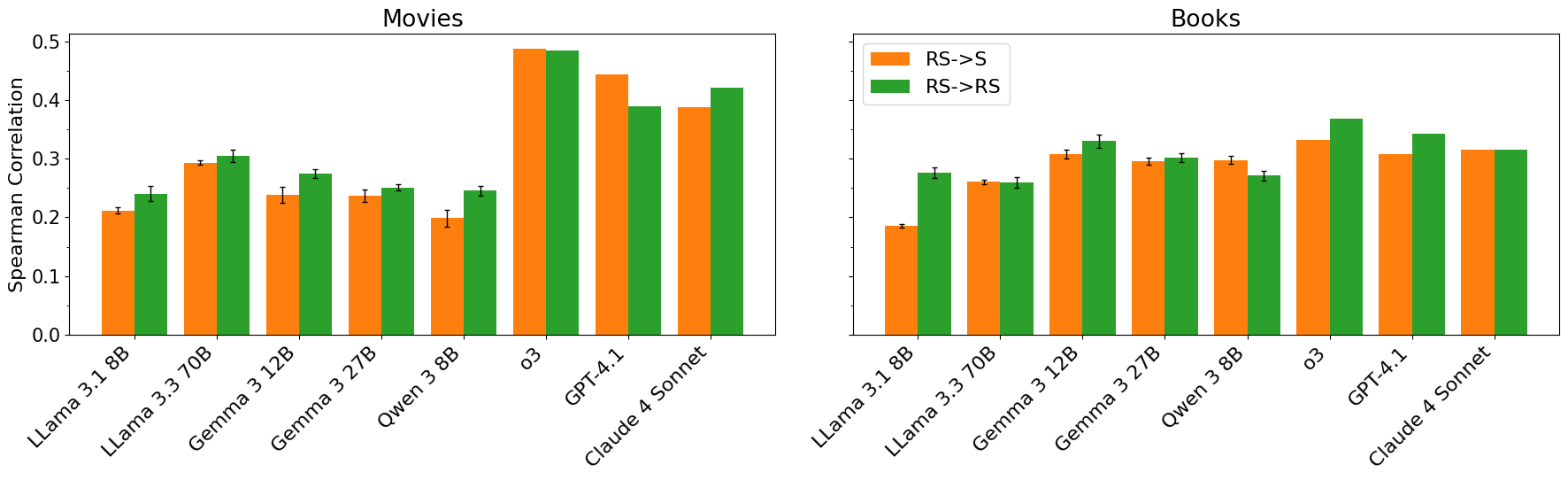}
  \caption{Average Spearman Correlation with \rstos and \rstors prompting.
  For the open-source models, error bars represent the standard deviation.
  \rstors improves the performance in general, and the effect is clearer for particular models such as Llama 3.1 8B or Gemma 3 12B.}
  \label{fig:review-writing-results}
\end{figure*}

\begin{figure}[ht]
  \centering
  \begin{subfigure}[b]{0.4\linewidth}
    \includegraphics[width=\linewidth]{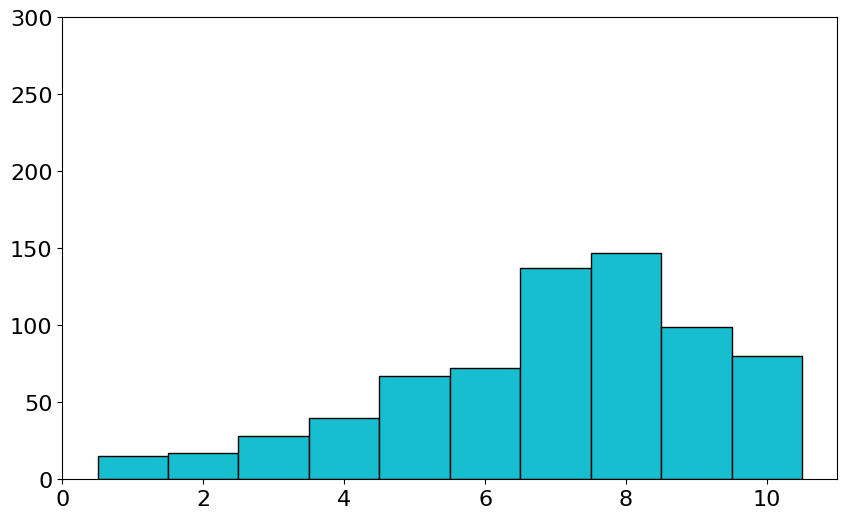}
    \caption{Ground Truth}
    \label{fig:gemma-ground-truth-distr}
  \end{subfigure}%
  \quad
  \begin{subfigure}[b]{0.4\linewidth}
    \includegraphics[width=\linewidth]{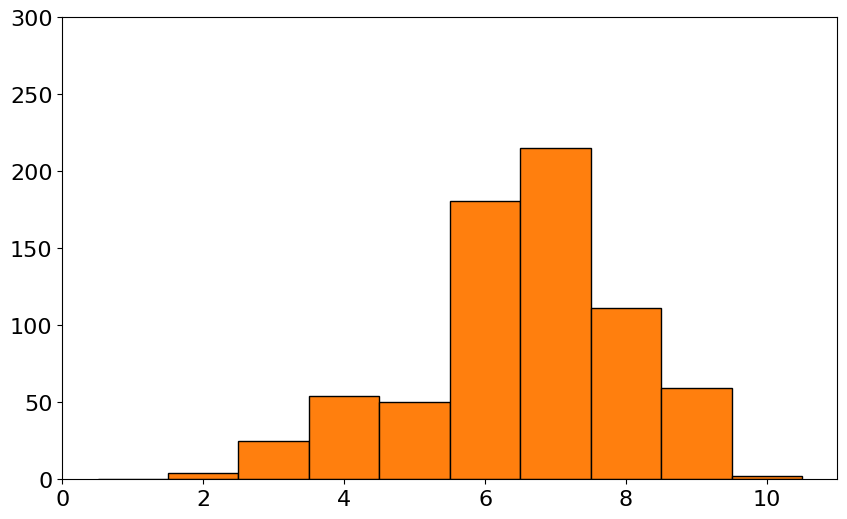}
    \caption{\rstos}
    \label{fig:gemma-rss-distr}
  \end{subfigure}

  \vspace{0.3em}

  \begin{subfigure}[b]{0.4\linewidth}
    \includegraphics[width=\linewidth]{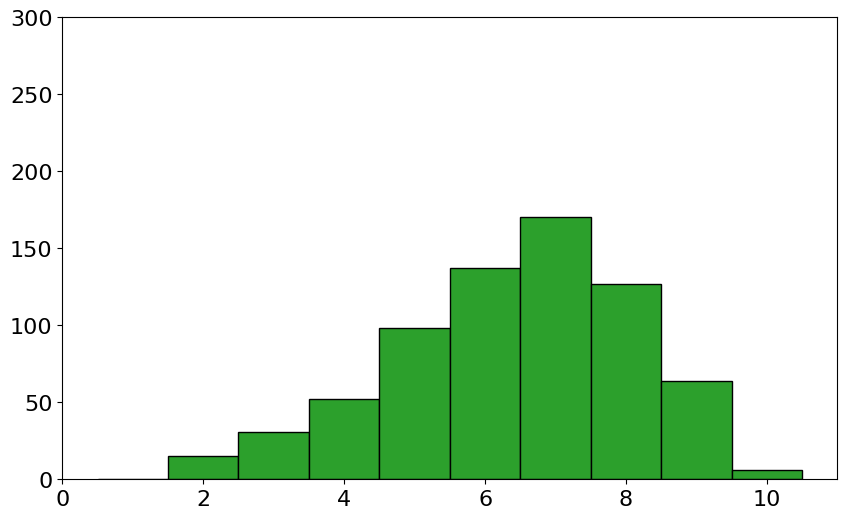}
    \caption{\rstors}
    \label{fig:gemma-rsrs-distr}
  \end{subfigure}%
  \quad
  \begin{subfigure}[b]{0.4\linewidth}
    \includegraphics[width=\linewidth]{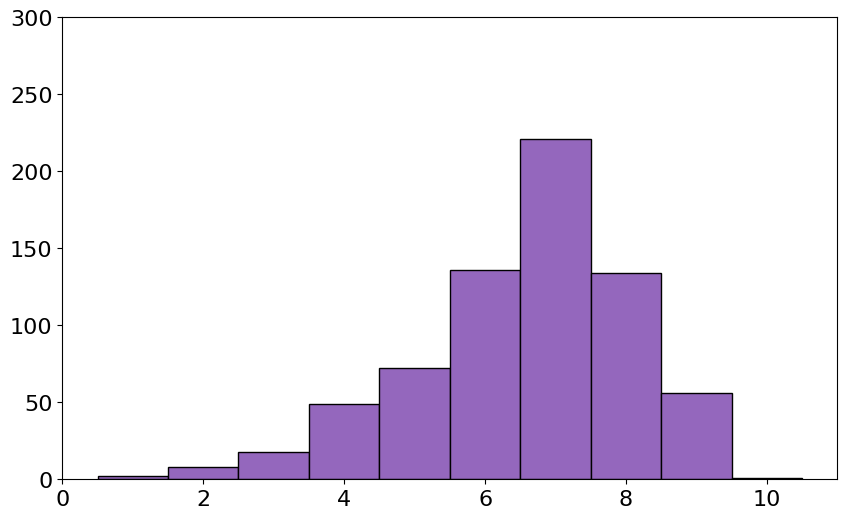}
    \caption{\rstors +  CoT}
    \label{fig:gemma-cot-distr}
  \end{subfigure}

  \caption{Output rating distribution of Gemma3 12B on the Movies dataset with different prompting methods. \rstors flatten the distribution compared to \rstosnospace, but adding CoT partially reverts the effect.}
  \label{fig:gemma-prompt-engineering-label-stats}
\end{figure}

\autoref{fig:self-description-results} compares the performance of the LLMs employing different prompting formats, with and without the self-described preference text.
The results clearly show that using the self-described preference (labeled \tos in the figure) yields significantly worse performance than using per-item reviews with scores (\rstos in the ``No Description'' groups).
This suggests that while the self-described preference may be enough for simpler tasks, more specific per-item review information is crucial for the more difficult rating prediction task. \par
When combining both data types, the results are mixed.
For the Movies dataset, providing the self-described preference along with the per-item review texts in \rstos prompt leads to a performance improvement for Gemma 3 12B.
However, for the other dataset and model combinations, the existence of the self-described preference degrades the performance.
A possible explanation for this phenomenon is the different nature of the review texts in the datasets.
The Movies dataset contains longer, more detailed review texts.
In this case, synthesized summaries may contain enough information for the models to improve the prediction.
On the other hand, the other datasets contain shorter reviews, which may not be informative enough when summarized.

\begin{figure*}[htb]
  \centering
    \includegraphics[width=\textwidth]{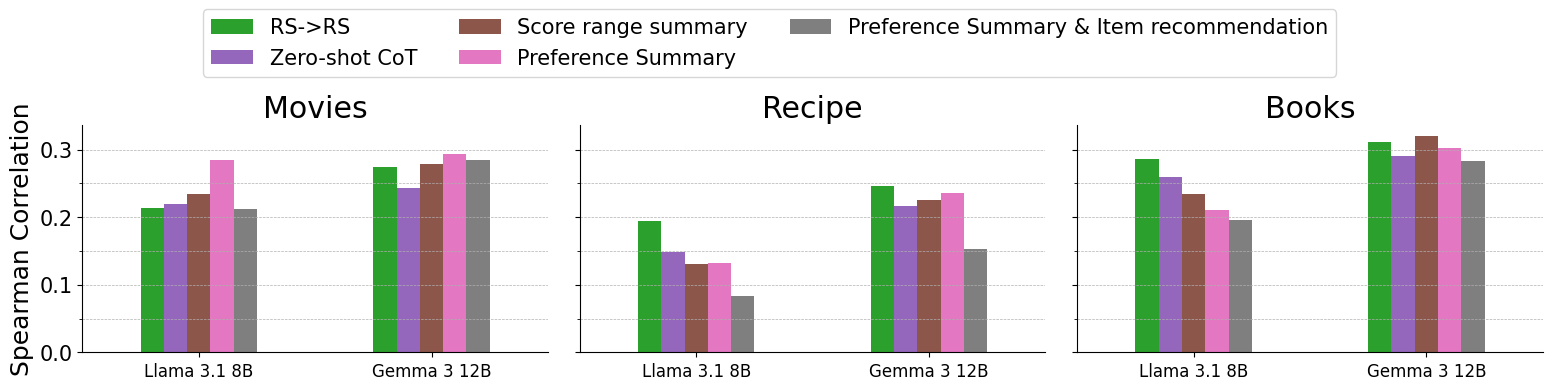}
  \caption{Comparison of the prompt engineering techniques on the user rating prediction task. In most cases the additional techniques do not result in the performance improvement compared to the original \rstors prompting.}
  \label{fig:prompt-engineering-results}
\end{figure*}

\section{Prompt Engineering}
\label{section:prompt-engineering}
In this section, we investigate whether the performance of review-based rating prediction can be further enhanced with prompt engineering. We first introduce a review-writing strategy as a natural extension of \rstos prompting. Then, we combine it with other techniques adapted from prior work to see if they can lead to even better performance.
\subsection{Review-Writing Prompt}
\label{subsection:review-writing}
First, we investigate a strategy where the LLM is prompted to generate a hypothetical review.
This approach is inspired by the PerSE framework for fine-tuned LLMs~\cite{permpst}.
We denote this format as Review+Score-to-Review+Score \textbf{(\rstorsnospace)}.
Formally, this involves modifying the instruction $I$ from the \rstos prompt to have the LLM output a tuple $(t'_u, y'_u)$, where $t'_u$ is the hypothetical review that the LLM expects user $u$ to write for the target item $x_u$. 
\subsection{Results of the Review-Writing}
\label{subsection:prompt-engineering-result}
We first analyze the effect of the review-writing (\rstorsnospace) prompt by comparing it against the baseline \rstos prompt. The full numerical results are included in Appendix \ref{appendix:prompt-engineering-results-detail}.\par
\autoref{fig:review-writing-results} shows that generating a hypothetical review generally improves Spearman correlation.
Across 24 model-dataset combinations, 15 show an improvement in mean correlation.
This improvement is statistically significant in 9 out of 15 combinations subjected to Welch's t-test.
The effect is particularly pronounced for smaller models such as Llama 3.1 8B and Gemma 3 12B, which show improvements across all three datasets.
\highlight
Notably, with the \rstors prompt, Gemma 3 12B even achieves better results than its larger variant, Gemma 27B, on all three datasets.
This result suggests that rating prediction conditioned on the personalized in-context data might be an exception to conventional scaling laws~\cite{scaling-law, chinchilla}, where larger models are typically assumed to be superior.
We leave a detailed analysis of this phenomenon for future work.
\finishhighlight
\par
However, the impact on RMSE is more limited. While a reduction in mean RMSE is observed in 17 of 24 combinations, the change is statistically significant in only 6 of 15 cases.
We even observe a trade-off for some models.
For instance, for Llama 3.1 8B and Gemma3 12B on the Movies dataset, while Spearman correlation improves, RMSE also increases.\par

To understand this trade-off, we analyze the output rating distributions.
We hypothesize that the review-writing process encourages the LLM to predict more extreme scores rather than always predicting neutral ratings.
\autoref{fig:gemma-prompt-engineering-label-stats} supports this hypothesis.
While the distribution for \rstos (\autoref{fig:gemma-rss-distr}) is heavily concentrated around neutral scores like six or seven, \rstors (\autoref{fig:gemma-rsrs-distr}) prompt produces a flatter distribution closer to the ground truth.
This wider range of outputs can improve correlation but may increase the absolute error for some predictions, thus explaining the trade-off.
Other models show slightly different trends as listed in Appendix \ref{appendix:output-distribution}.

\subsection{Combination with Other Strategies}
\label{subsection:other-prompt-engineering}
Next, we examine whether the performance can be further enhanced by combining \rstors prompt with other prompting strategies from related work.
Concrete prompts are provided in Appendix \ref{appendix:prompt-engineering}.
\paragraph{Zero-shot CoT}
Following \citet{zero-shot-cot}, we append ``Let's think step by step'' to the prompt to trigger the LLM's reasoning capability. 
\paragraph{Score Range Summary}
Adapted from \citet{intsum}, this prompt first instructs the LLM to first summarize the user's past rating range.
\paragraph{Preference Summary}
Inspired by KAR~\cite{kar}, this prompts the LLM to first summarize the user's preferences.
\paragraph{Preference Summary + Item Recommendation}
In addition to the above, this format, employing another prompt from LLM-Rec~\cite{llm-rec}, asks for both a preference summary and a recommendation justification before the final prediction.

\subsection{Results of Combined Strategies}
\label{prompt-engineering}
\autoref{fig:prompt-engineering-results} shows the results on Llama 3.1 8B and Gemma 3 12B models.
No additional technique consistently outperforms the \rstors baseline.
In particular, zero-shot CoT prompting leads to worse performance in five out of six cases.
This finding supports previous work~\cite{not-cot}, which suggests that general zero-shot CoT is less effective for tasks that do not require logical reasoning.
The output distribution in \autoref{fig:gemma-prompt-engineering-label-stats} illustrates the possible cause of that difference. 
Although \rstors (\autoref{fig:gemma-rsrs-distr}) flattens the distribution, applying CoT on it (\autoref{fig:gemma-cot-distr}) reverses this effect, resulting in the more frequent prediction of the neutral scores again. 
With zero-shot CoT, LLMs tend to output the analysis results for both likes and dislikes of the users at the same time, which may result in the ``balanced'' output score. 
Concrete generation examples are available in Appendix \ref{appendix-prompt-engineering-results}.

\section{Conclusion}
In this work, we conducted a comprehensive investigation into the performance of off-the-shelf LLMs on rating prediction, providing user-written review texts as in-context preference information.
Our findings demonstrated that the review texts significantly and consistently improve the performance across different models and datasets, and the models achieve results comparable to traditional baseline methods.
Notably, this improvement with review texts was observed even when the in-context items were not similar to the target item.
\par

Moreover, our comparative analysis revealed that the per-item review texts are more effective than the self-described preference data used in prior work for simpler tasks.\par
We also found that further performance enhancement can be achieved by instructing LLMs to generate a hypothetical review before predicting the ratings.
We provided evidence suggesting that the output rating distribution shift is one reason for this phenomenon.
\par
Our results highlight the potential of off-the-shelf LLMs as lightweight recommendation systems, potentially mitigating the cold-start problem.
This work also confirms the value of user-written reviews as a rich data source for personalization.
We hope these findings will lead to a future implementation of data-efficient personalization systems based on off-the-shelf LLM.

\newpage
\section*{Limitations}
Our model configurations are non-deterministic, so the results may differ with different random seeds. Moreover, excluding failed examples may have inappropriate effects when evaluating correlation metrics, and this exclusion may have unexpectedly affected the results.\par
Another limitation concerns the data contamination. As the knowledge cutoff for some of the tested models happened after the datasets were published, memorization of the data may affect the result. We need further investigation about how large the effect is.\par
\highlight
Our comparative analysis did not include methods based on fine-tuned models, which serve as important baselines. Furthermore, we did not evaluate the sensitivity of our approach to small variations in prompt phrasing. Addressing these two points is crucial for future work to properly position our method, which relies on off-the-shelf LLMs, within the broader landscape of LLM-based recommendation systems.\par
\finishhighlight
Finally, the analysis of self-described preferences relies on text transformation performed by LLMs, which may affect the quality of the generated preference texts.
Although we manually check the similarity of the generated texts with the examples used in previous studies, it is still possible that the artificially generated preference texts have qualitative differences from human-written texts.
Again, a new dataset with different styles of preference text from the same user is needed for a more accurate comparison.

\section*{Ethical Considerations}
The three datasets used in our study are based on user-generated contents crawled from online services. None of the datasets contains sensitive user information, and we ensure we do not disclose any personally identifiable information as part of our work.\par
In addition, providing the user information in the context of deployed LLM-based systems might result in an unexpected information leakage. Although our work expects the situation where only the data obtained from the target user is used, developers need to pay attention to handling sensitive data when implementing a similar system.\par
\highlight
Finally, our proposed method, which relies on review texts, may introduce several types of bias.
Our requirement for in-context examples with relatively long review texts means that our dataset may not be representative of the broader user population. 
Furthermore, by focusing on items with sufficient review data, our method could unintentionally favor popular items and underrepresent niche items, thus reducing the diversity of recommendations.
A comprehensive investigation of those risks is required in future work.
\section*{Acknowledgments}
We thank the three anonymous reviewers for their helpful comments and feedback.
This work was supported by JSPS KAKENHI Grant Number JP24H00809, Japan.
\finishhighlight

\bibliography{custom.bib}

\appendix
\section{Difficult Settings}
\label{rq-2}
In this section, we verify how the review-based rating prediction with off-the-shelf LLM works in situations with more limited resources.
\subsection{Variants of In-Context Examples}
\label{in-context-example-variants}
We make the preference prediction problem more difficult by providing the in-context preference information in the following ways.

\paragraph{Fewer}
First, we investigate the effect of the number of in-context examples.
With the same datasets introduced in \autoref{base-datasets}, we reduce the number of in-context examples to $k = 1, 3$, and compare the results with \autoref{subsection:base-prompting-results}, which uses $k = 5$.

\paragraph{Shorter}
Second, we examine the performance change in the situation where each review is a shorter text.
We create the Books (Short) dataset by sampling reviews with fewer than 200 characters from the same Amazon Reviews'23~\cite{amazon-books}, which is also used for the standard Books dataset. 
To exclude extremely short reviews, such as single words, we also set a length of 10 as the lower threshold. 
See Appendix \ref{appendix:dataset-stats} for more detailed statistics of the dataset.\par

\paragraph{Shuffle}
Third, we randomly shuffle the in-context review texts to verify whether LLMs improve user rating prediction performance by identifying target user characteristics from user review contents.\par
We create the Movies (Shuffle) dataset, which is made by shuffling the in-context examples of the Movies dataset in \autoref{base-datasets}. Therefore, in \rstors and \rstos settings, the target user's past review scores are paired with unrelated reviews written by other users. 

\subsection{Results and Analysis}
\begin{figure}[t]
  \centering
    \includegraphics[width=1.0\columnwidth]{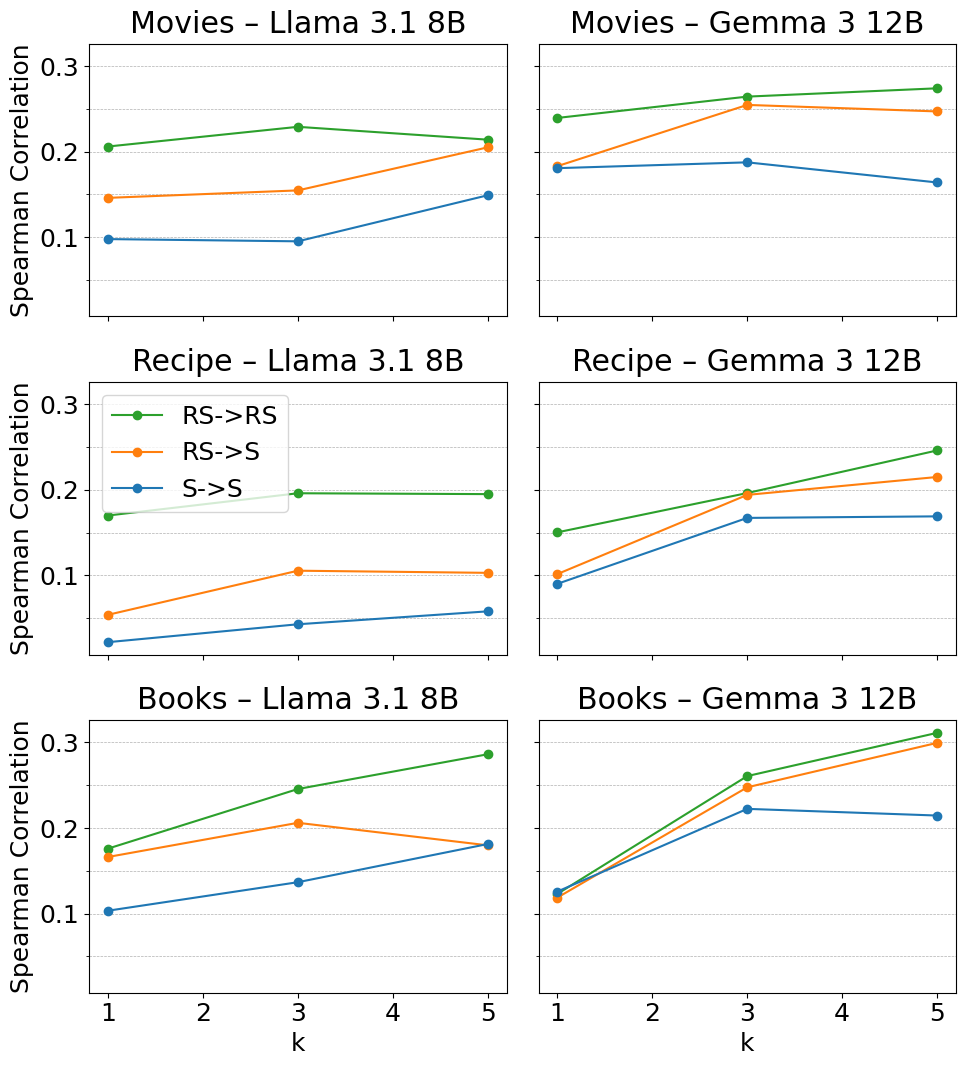}
  \caption{Comparison of the results with $k=1, 3, 5$. \rstors and \rstos enhance the performance even with a fewer in-context examples.}
  \label{fig:few-shot-results}
\end{figure}

\begin{figure}[t]
  \centering
    \includegraphics[width=1.0\columnwidth]{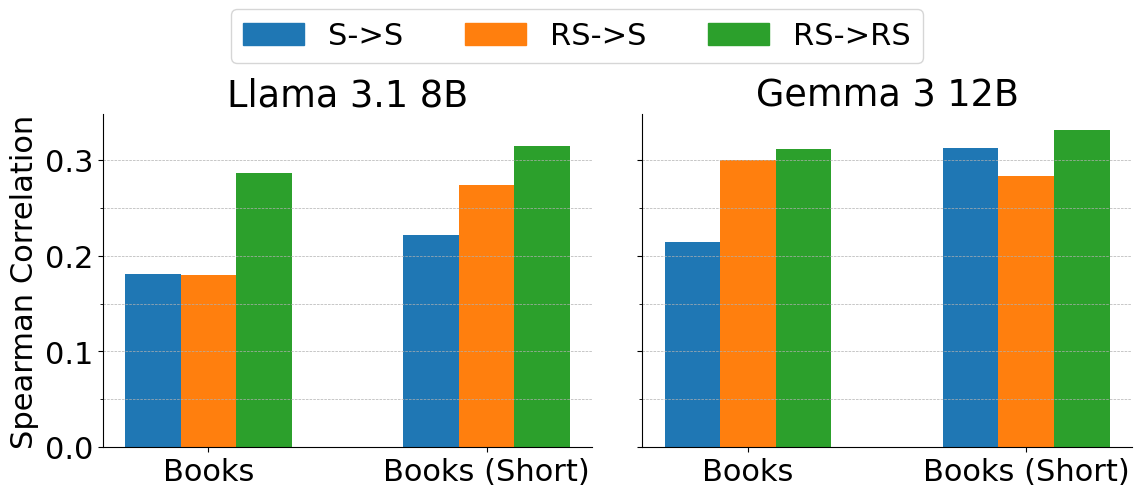}
  \caption{Comparison of the results with the Books and the Books (Short) datasets. Shorter reviews still lead to the performance improvement.}
  \label{fig:short-results}
\end{figure}

\begin{figure}[t]
  \centering
    \includegraphics[width=1.0\columnwidth]{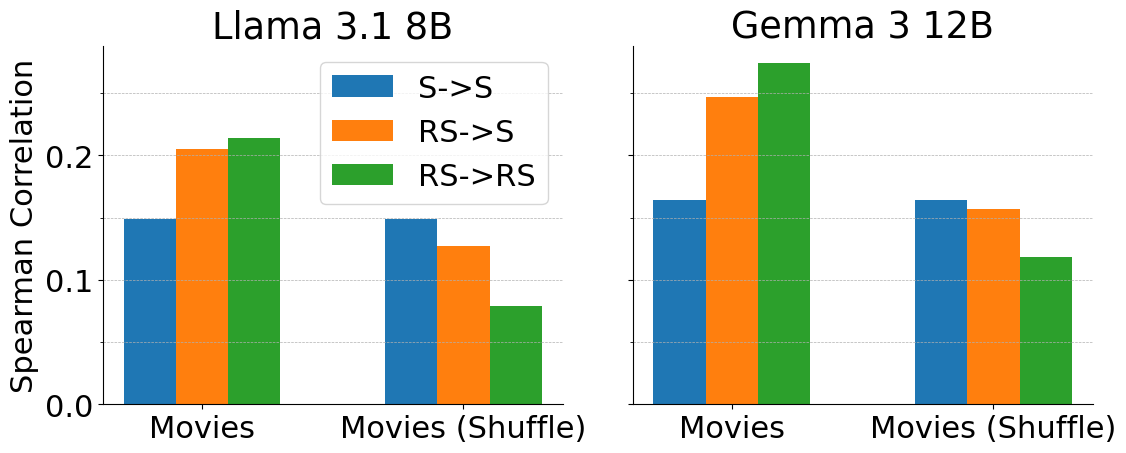}
  \caption{Comparison of the results with the Movies and the Movies (Shuffle) datasets. \rstos and \rstors prompting worsen the performance on the Movies (Shuffle) dataset, which suggests that the LLMs actually reference the review contents to predict the target user's preference.}
  \label{fig:shuffle-results}
\end{figure}

Figure \ref{fig:few-shot-results}, \ref{fig:short-results}, and \ref{fig:shuffle-results} show the results on the two settings with Llama 3.1 8B and Gemma 3 12B.
Both models perform better with \rstors and \rstos compared to \stosnospace, even with fewer in-context examples such as $k = 1, 3$.
\rstors also marks higher performance than \rstosnospace. 
The results suggest that the findings in \autoref{subsection:base-prompting-results} still hold with an extremely small number of in-context examples.\par
The short review experiment also supports a similar conclusion.
Both LLMs show improved performance on the Books (Short) dataset with \rstors compared to \stosnospace. 
This suggests that even short review texts can contribute to the rating prediction task performed by off-the-shelf LLMs.\par
Although the degree of improvement looks smaller than that with the standard Books dataset, direct comparison is not appropriate because of the difference in rating prediction difficulty in both datasets. 
As shown in Appendix \ref{appendix:dataset-stats}, users extracted for the Books (Short) dataset show smaller variance in their integer preference scores, which makes it easier to predict the scores in the Books (Short) dataset solely from the numeric ratings. 
We leave a more rigorous comparison for future work. \par
Performance improvement by using review texts cannot be observed in the shuffle setting. 
On the Movies (Shuffle) dataset, since the review texts are more incorporated into the prediction process in \rstos and \rstors prompting settings, a significant drop in the prediction performance is observed for the Shuffle dataset, contrary to the improvement in the standard dataset. 
This result indicates that the LLMs actually reference the review contents to predict the target user's preference, which means that giving the correct reviews as in-context examples is at least required for performance enhancement.

\section{Verification of the main results}
\highlight
\subsection{Variance of the predictions across multiple runs}
\label{appendix:variance-multiple-runs}
\begin{figure*}[!t]
  \centering
    \includegraphics[width=\textwidth]{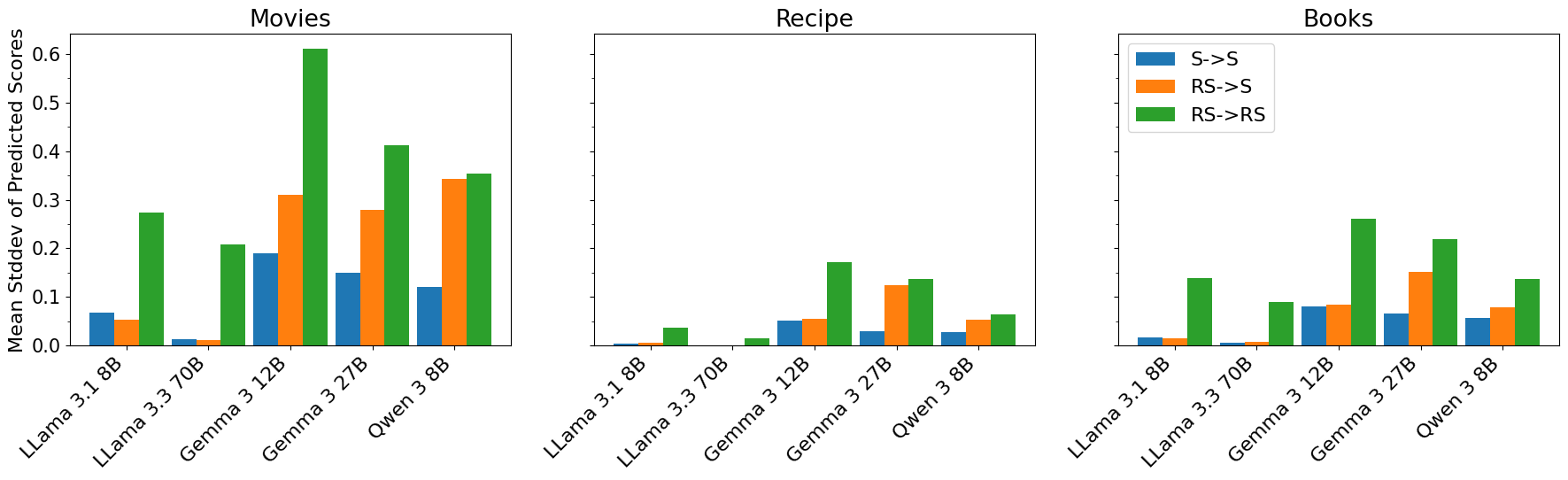}
  \caption{Average standard deviation of the predicted values per examplee across multiple runs.}
  \label{fig:predicted-scores-stddev}
\end{figure*}

Our experimental settings involve stochastic sampling, which can introduce variance into the results.
To assess the stability of our findings, we analyze the variance in predictions across multiple runs in \autoref{subsection:base-prompting-results}  and \autoref{subsection:prompt-engineering-result}.
Specifically, we calculate the average standard deviation of the predicted scores for each test instance over the six experimental runs for the open-source models.\par
The results are shown in \autoref{fig:predicted-scores-stddev}. Note that different series of models use different sampling parameters.
The variance of predictions differs across model series, partly due to the use of different sampling parameters that were optimized for each model family.
In particular, for the Llama series models, we set the temperature to a very low value $(t = 0.01)$ to mitigate generation failures (e.g., parsing errors) in the more complex \rstors settings.\par
For different models and datasets, the prediction variance tends to increase as the prompting format changes from \stosnospace, to \rstosnospace, and finally to \rstorsnospace.
A controlled comparison of different models under identical sampling parameters is left as future work.

\subsection{Analysis of Extrapolation data points}
\label{appendix:extrapolation}
\begin{table*}[ht]
  \centering
  \resizebox{\textwidth}{!}{%
  \begin{tabular}{@{}ll
                  rrr  %
                  rrr  %
                  rrr@{}} %
    \toprule
    Dataset & Model
    & Ground Truth
    & Prediction (S $\to$ S)
    & Prediction (RS $\to$ S)
    & Prediction (RS $\to$ RS)
    \\
    \midrule
Movies & LLama 3.1 8B
	& 183 & 164.333 & 91.833 & 59.000
\\
 & LLama 3.3 70B
	&  & 53.833 & 70.500 & 84.500
\\
 & Gemma 3 12B
	&  & 34.333 & 60.333 & 76.000
\\
 & Gemma 3 27B
	&  & 19.833 & 68.167 & 50.667
\\
 & Qwen 3 8B
	&  & 23.000 & 59.000 & 36.333
\\
 & o3
	&  & 73.000 & 59.000 & 45.000
\\
 & GPT-4.1
	&  & 28.000 & 44.000 & 27.000
\\
 & Claude 4 Sonnet
	&  & 96.000 & 97.000 & 98.000
\\
\midrule
Recipe & LLama 3.1 8B
	& 73 & 19.333 & 10.333 & 6.667
\\
 & LLama 3.3 70B
	&  & 1.000 & 1.000 & 2.000
\\
 & Gemma 3 12B
	&  & 26.833 & 42.500 & 32.167
\\
 & Gemma 3 27B
	&  & 6.000 & 39.500 & 9.000
\\
 & Qwen 3 8B
	&  & 0.333 & 6.333 & 1.667
\\
 & o3
	&  & 12.000 & 6.000 & 0.000
\\
 & GPT-4.1
	&  & 7.000 & 6.000 & 0.000
\\
 & Claude 4 Sonnet
	&  & 44.000 & 6.000 & 2.000
\\
\midrule
Books & LLama 3.1 8B
	& 106 & 32.000 & 20.500 & 14.667
\\
 & LLama 3.3 70B
	&  & 13.833 & 40.833 & 15.667
\\
 & Gemma 3 12B
	&  & 54.667 & 67.000 & 55.833
\\
 & Gemma 3 27B
	&  & 35.500 & 59.500 & 37.000
\\
 & Qwen 3 8B
	&  & 5.833 & 25.833 & 13.833
\\
 & o3
	&  & 33.000 & 28.000 & 5.000
\\
 & GPT-4.1
	&  & 33.000 & 50.000 & 7.000
\\
 & Claude 4 Sonnet
	&  & 119.000 & 66.000 & 38.000
\\
\midrule
  \end{tabular}%
  }
\caption{Number of extrapoation class values in the groud truth labels and the predictions with different models and the prompting methods.}
  \label{tab:extrapolation-count}
\end{table*}

\begin{figure*}[!t]
  \centering
    \includegraphics[width=\textwidth]{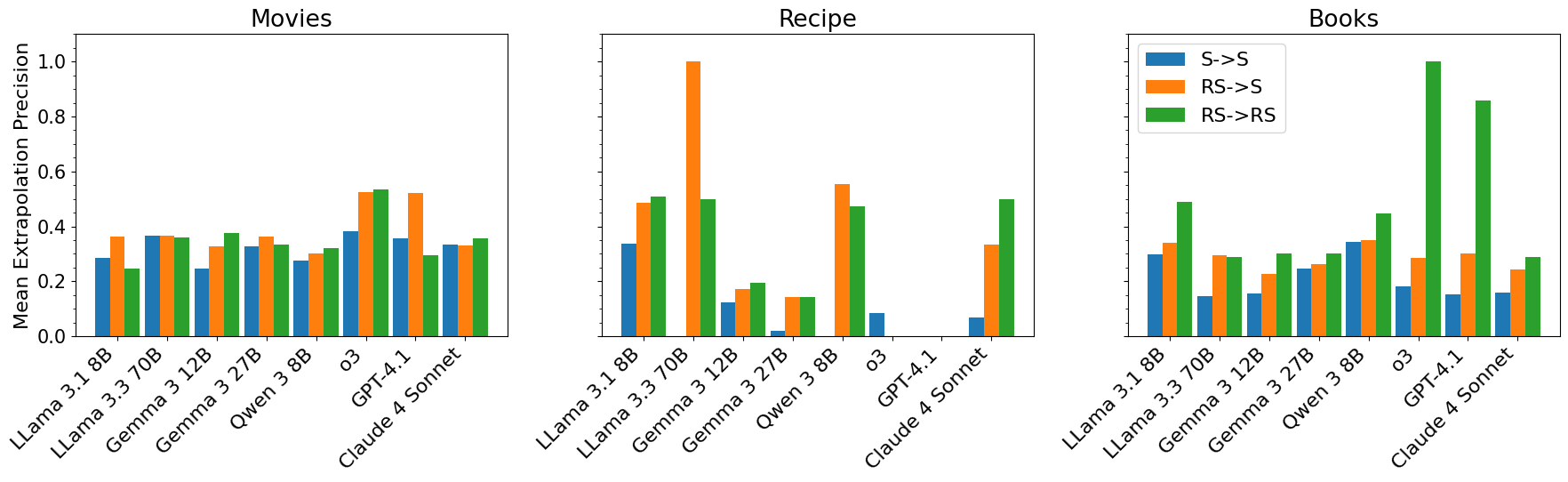}
    \includegraphics[width=\textwidth]{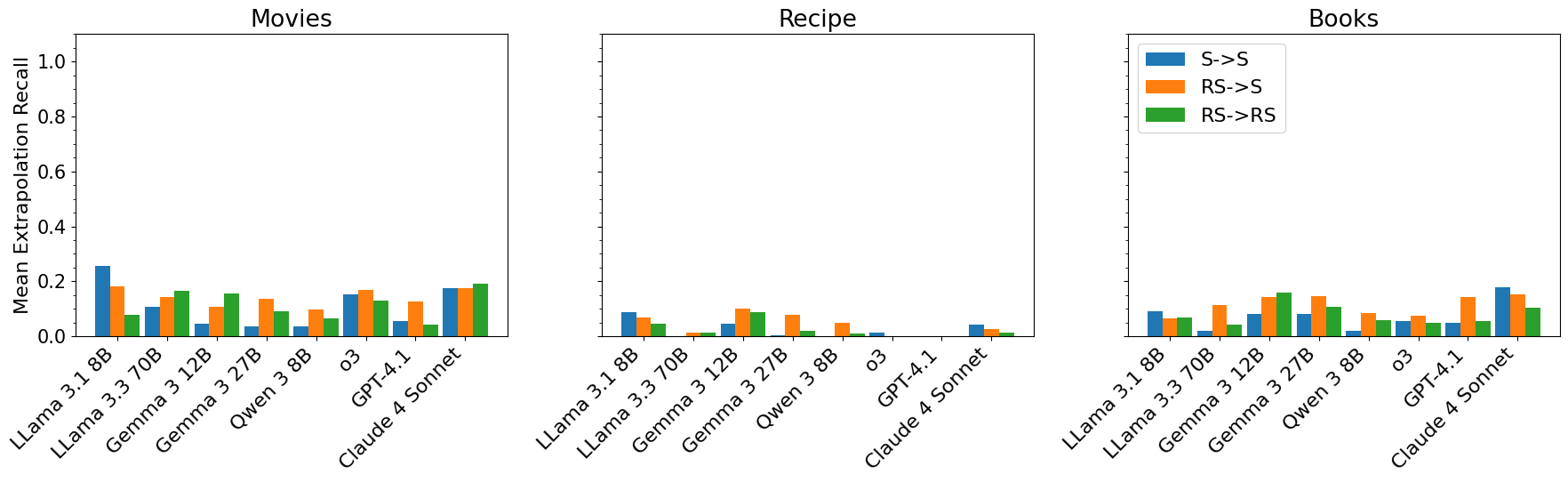}
  \caption{Average extrapolation precision (top) and recall (bottom) with different prompting format.}
  \label{fig:extrapolation-results}
\end{figure*}

Unlike traditional methods such as MF and User Average, which are inherently constrained by the range of observed ratings, LLMs can theoretically predict scores outside the range of in-context examples.
To verify this extrapolation capability, we re-analyze the results in  \autoref{subsection:base-prompting-results}  and \autoref{subsection:prompt-engineering-result}.\par
We reframe the task as a ternary classification problem based on the score range of the in-context examples.
For a given set of in-context scores, let $c_{min}$, $c_{max}$ be the minimum and maximum values
We then classify both the ground-truth score $y_u$ and the predicted score $y_u'$ into one of the three labels:
Any ground truth labels and predicted scores with value $v$ associated with those in-context examples are treated as one of the following labels.
\begin{itemize}
    \setlength{\parskip}{0cm}
    \setlength{\itemsep}{0cm}
    \item $l_{low}$: if $v < c_{min}$
    \item $l_{in}$: if $c_{min} \leq v \leq c_{max}$
    \item $l_{high}$: if $c_{max} < v$
\end{itemize}
We call $l_{low}$ and $l_{high}$ as ``extrapolation classes''.
We then measure ``extrapolation precision'' and ``extrapolation recall,'' which are the micro-averaged precision and recall over these two extrapolation classes.\par

\autoref{tab:extrapolation-count} shows the number of extrapolation class values in the ground truth and the model predictions.
As shown, all three datasets contain a non-trivial number of extrapolation class ground truth labels, and most models are capable of making such extrapolated predictions across different datasets and prompting methods.\par
\autoref{fig:extrapolation-results} presents the extrapolation precision and recall.
While recall is generally low across all datasets, the precision is often reasonable.
The benefit of richer prompts is particularly evident for the Books dataset. 
Here, both adding review texts (\stos to \rstosnospace) and prompting for hypothetical review texts (\rstos to \rstorsnospace) improve extrapolation precision.
This improvement occurs even as some models increase their volume of extrapolated predictions. 
This ability to predict unseen rating levels represents a potential advantage of LLM-based rating prediction over traditional algorithms.

\finishhighlight
\subsection{Self-described Preference Generated by Different Models}
\label{appendix-self-description-generation}
\begin{figure}[t]
  \centering
    \includegraphics[width=1.0\columnwidth]{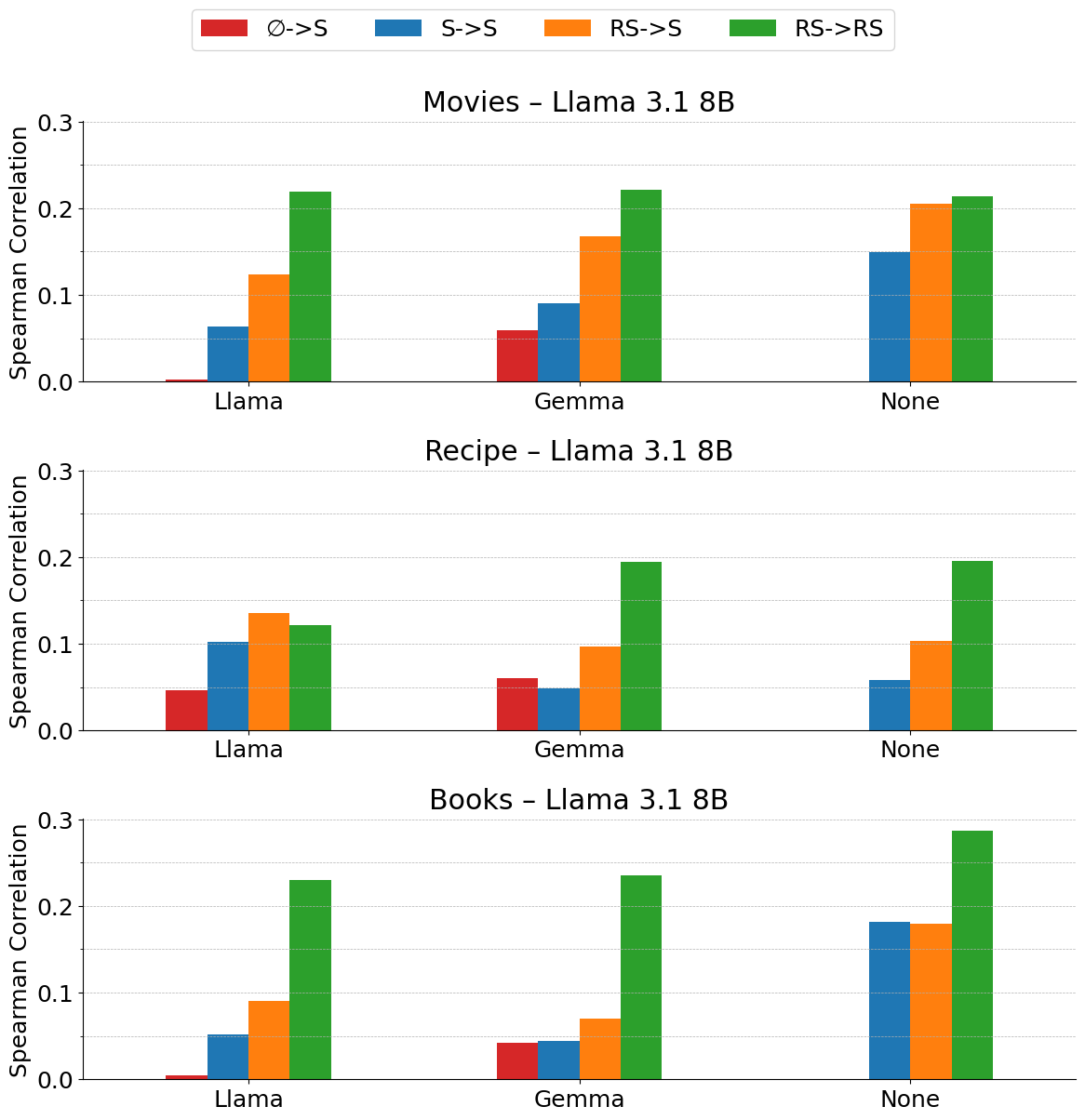}
  \caption{Comparison Llama 3.1 8B's performance with self-description generated by different LLMs}
  \label{fig:description-generator-comparison}
\end{figure}

In \autoref{self-description}, we only use the self-described preference generated by the same model as the one that performs the preference prediction. To check whether the quality of the text transformation affects the result of the user rating prediction, we repeat the same experiment as \autoref{self-description} with Gemma 3 12B as the self-described preference generator and Llama 3.1 8B as the user rating predictor.\par
\autoref{fig:description-generator-comparison} reports the result. Although the performance with \tos is slightly improved when Gemma 3 12B is used as the self-described preference generator model, it is still worse than the \rstos without the self-description text. This result suggests that the impact of the model selection on self-described preference generation is lower than that of the existence of the per-item review texts.

\section{Implementation Details}
\subsection{Models}
\label{appendix-models}
During inference with open-source models, we limit the maximum number of generated tokens to 768 for the Llama and Gemma models. For Qwen 3 8B, we set this to 32768 to allow more extended reasoning.\par
We set the temperature to 0.01 for the Llama models. Other parameters follow the default set on the huggingface pages\footnote{https://huggingface.co/meta-llama/Llama-3.1-8B-Instruct}\footnote{https://huggingface.co/meta-llama/Llama-3.3-70B-Instruct}\footnote{https://huggingface.co/google/gemma-3-12b-it}\footnote{https://huggingface.co/google/gemma-3-27b-it}\footnote{https://huggingface.co/Qwen/Qwen3-8B} as of 2025 July.\par

For inference with o3, GPT-4.1, and Claude 4 Sonnet, we used o3-2025-04-16, gpt-4.1-2025-04-14, and claude-sonnet-4-20250514 snapshots, respectively. For other parameters,  the default values set with LiteLLM\footnote{https://github.com/BerriAI/litellm} are used.

\subsection{Computational Resources}
We conducted the experiments with different numbers of NVIDIA A100 (40GB), depending on the LLM used for each run. We report the number of GPUs used and the maximum hours spent for each run in \autoref{subsection:base-prompting-results} with each model as follows:

\begin{itemize}
    \setlength{\parskip}{0cm}
    \setlength{\itemsep}{0cm}
    \item Llama 3.1 8B: 1 GPU, 2 hours
    \item Llama 3.3 70B: 4 GPUs, 6 hours 
    \item Gemma 3 12B: 1 GPU, 4 hours
    \item Gemma 3 27B: 2 GPUs, 6 hours
    
    \item Qwen 8B: 1 GPUs, 4 hours
    
\end{itemize}

Each run in \autoref{rq-2}, \autoref{section:prompt-engineering}, and \autoref{section:self-described-preference} took the same number of GPUs and twice as much time as listed above because of the required intermediate outputs. 

\subsection{Dataset Statistics}
\label{appendix:dataset-stats}

\begin{table*}[ht]
  \centering
    \resizebox{\textwidth}{!}{%
    \begin{tabular}{@{}lrrrr@{}}
      \toprule
      Dataset         & Num of Examples & Avg Item Description Length & Avg Review Length & Avg Per-user Score Stddev \\
      \midrule
      Movies          & 702              & 1142.0                      &  752.8             & 1.54                      \\
      Recipe          & 1000             &  766.8                      &  370.6             & 0.44                      \\
      Books           & 1000             & 1134.2                      &  650.7             & 0.74                      \\
      Books (Short)   & 1000             & 1075.2                      &   84.8             & 0.49                      \\
      \bottomrule
    \end{tabular}%
    }
  \caption{Dataset‐level statistics: number of examples, average item‐description length (characters), average review length (characters), and per‐user score standard deviation.}
  \label{tab:dataset_stats}
\end{table*}

\begin{table}[ht]
  \centering
    \resizebox{\linewidth}{!}{%
    \begin{tabular}{@{}lrr@{}}
      \toprule
      Dataset         & Avg. Total Reviews per User & Score Representativeness (Correlation)\\
      \midrule
      Movies          & 28.8 & 0.942 \\
      Recipe          & 74.2 & 0.593 \\
      Books           & 34.5 & 0.813 \\
      \bottomrule
    \end{tabular}%
    }
  \caption{
    Analysis of the users sampled for our experiments and the representativeness of their in-context scores. 
    The 'Avg. Total Reviews per User' column shows the average number of reviews these users wrote in the original, full dataset. 
    The 'Score Representativeness (Correlation)' column measures the Spearman correlation between a user's overall average score (from all their reviews) and the average of the five scores specifically chosen as their in-context examples. 
    Note that the Movies dataset was used in its entirety without user sampling.
  }
  \label{tab:dataset-incontext-scores-stats}
\end{table}

\begin{figure}[ht]
  \centering
  \begin{subfigure}[b]{0.48\linewidth}
    \includegraphics[width=\linewidth]{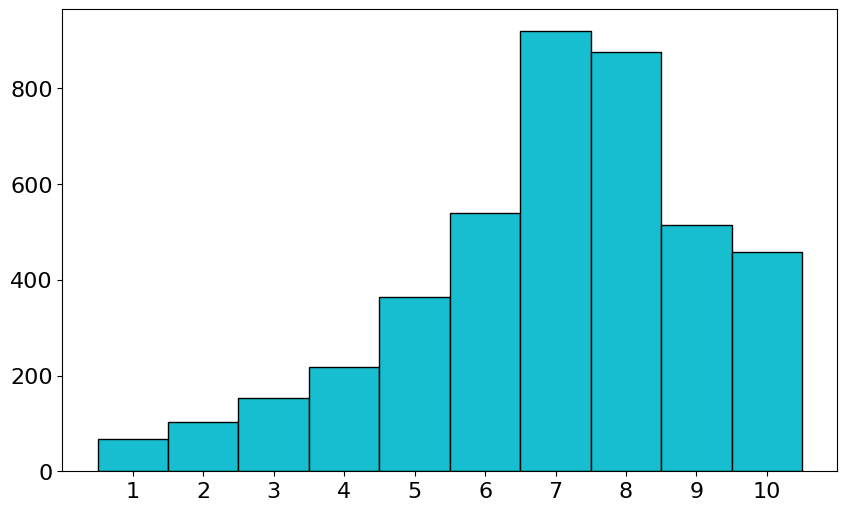}
    \caption{Movies}
    \label{fig:movies-labels}
  \end{subfigure}%
  \hfill
  \begin{subfigure}[b]{0.48\linewidth}
    \includegraphics[width=\linewidth]{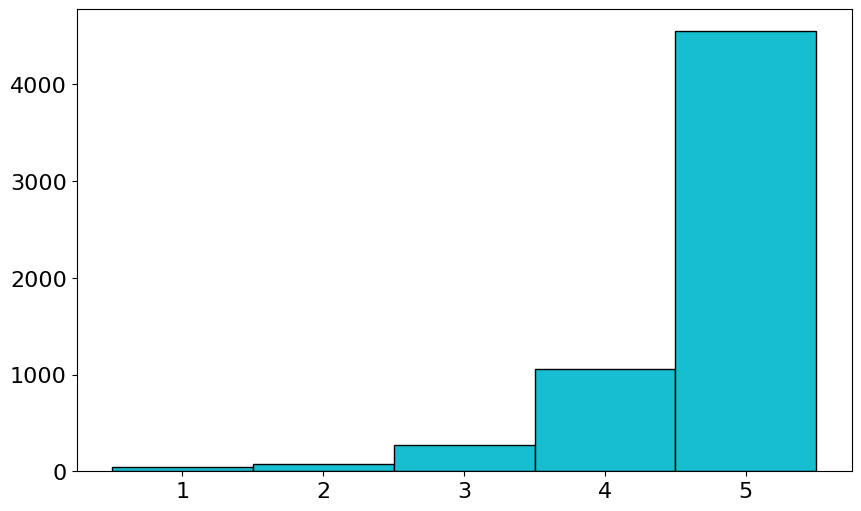}
    \caption{Recipe}
    \label{fig:recipe-labels}
  \end{subfigure}

  \vspace{1em}

  \begin{subfigure}[b]{0.48\linewidth}
    \includegraphics[width=\linewidth]{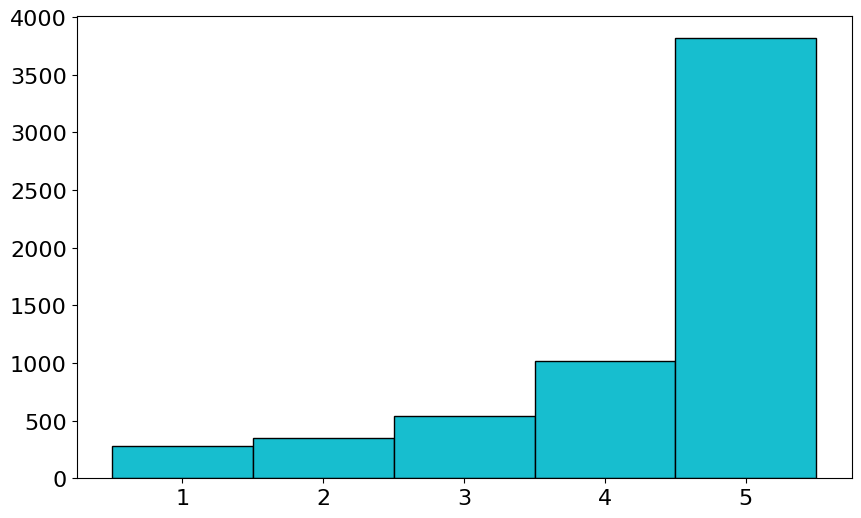}
    \caption{Books}
    \label{fig:books-labels}
  \end{subfigure}%
  \hfill
  \begin{subfigure}[b]{0.48\linewidth}
    \includegraphics[width=\linewidth]{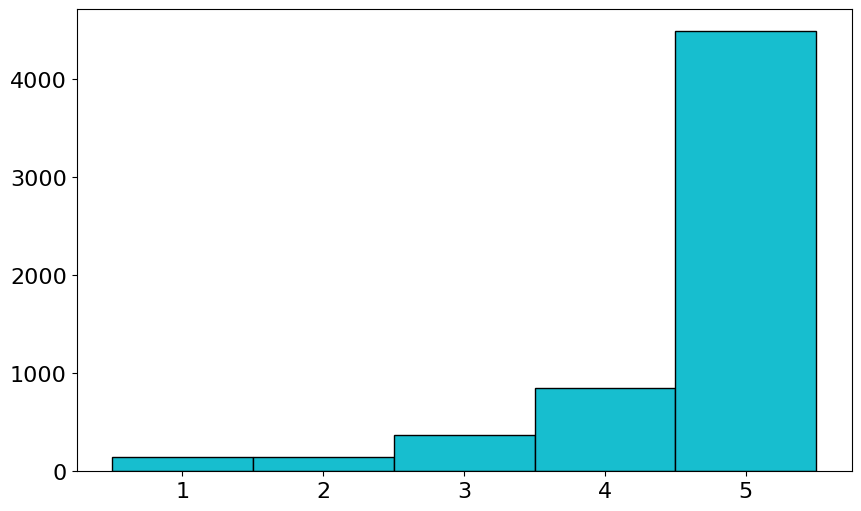}
    \caption{Books (Short)}
    \label{fig:books-short-labels}
  \end{subfigure}

  \caption{Label distribution of each dataset (including the in-context examples)}
  \label{fig:label-stats}
\end{figure}

We show the statistics about the datasets we used in the experiments in \autoref{tab:dataset_stats}. We also present the numeric score distribution in \autoref{fig:label-stats}. Note that for the Movies (Shuffle) dataset, all the values are the same as those of the standard Movies dataset, since the dataset is just made by shuffling the review text data in the original dataset.\par
\highlight
Since the Books and Recipe datasets were generated by sampling from their larger, original versions, we also analyzed the representativeness of our sampling method.
\autoref{tab:dataset-incontext-scores-stats} presents this analysis, showing the average total number of reviews per user alongside the Spearman correlation between a user's overall average score (from the original dataset) and the average of the five scores sampled for their in-context examples.
As the low correlation for the Recipe dataset indicates, the scores in our sampled in-context examples are less representative of the users' general scoring tendencies.
This difference is likely due to the two factors.
First, the dataset has a highly skewed score distribution as seen in \autoref{fig:label-stats}.
Second, our sampling strategy prioritizes reviews that meet a minimum length threshold.
However, these two conditions also apply to the Books dataset.
We need to investigate further to differentiate between those two datasets.
\finishhighlight

\subsection{Other Software and Artifacts}
\label{appendix-other-software}
We ran the code for all the experiments with Python 3.11.10.
For open-source LLM inference, we used PyTorch~\cite{pytorch} 2.6.0 and 
Transformers~\cite{transformers} 4.50.0. For closed LLMs, we used LiteLLM\footnote{https://github.com/BerriAI/litellm} 1.74.0. We calculated the evaluation metrics with scikit-learn~\cite{sklearn} 1.6.1 and 
SciPy~\cite{scipy} 1.15.1.

\subsection{\rstorsnospace, \rstos and \stos Prompts}
\label{appendix:prompts}
\begin{figure}
\begin{Prompt}

\begin{verbatim}
<|start_header_id|>system<|end_header_id|>
You function as an insightful assistant whose
role is to assist individuals in making
decisions that align with their personal
preferences. Use your understanding of their
likes, dislikes, and inclinations to provide 
relevant and thoughtful recommendations.
<|eot_id|>

<|start_header_id|>user<|end_header_id|>
[User Question] You will be presented with
several plot summaries, each accompanied by
a review from the same critic. Your task is
to analyze both the plot summaries and the
corresponding reviews to discern the
reviewer's preferences. Afterward, consider
a new plot and create a review that you
believe this reviewer would write based on
the established preferences. 

{icl_example}

Please follow the above critic and give a
review for the given plot. Your response
should strictly follow the format: 
```json
{{
  "Review": "<proposed review conforms to
             style demonstrated in the previous
             reviews>",
  "Score": <1-10, 1 is the lowest and
            10 is the highest>
}}
```
Please remember to replace the placeholder
text within the "<>" with the appropriate
details of your response.

[The Start of Plot]
{plot}
[The End of Plot]
<|eot_id|>

<|start_header_id|>assistant<|end_header_id|>
[Review] Here is the Json format of the review: 
\end{verbatim}
\end{Prompt}
\caption{Query Prompt used for RS $\to$ RS examples}
\label{prompt:base-query-prompt}
\end{figure}

\begin{figure}
\begin{Prompt}
\begin{verbatim}
[The Start of Plot {n}]
{plot}
[The End of Plot {n}]
[Review]
```json
{{
  "Review": "{review}",
  "Score": {score}
}}
```
\end{verbatim}
\end{Prompt}
\caption{In-Context Example Template used for RS $\to$ RS examples}
\label{prompt:base-ice-template}
\end{figure}

We present the base prompt used for Llama Models and Movies dataset with \rstors settings in \autoref{prompt:base-query-prompt}. The prompt is adopted from PerSE~\cite{permpst}. The "\{plot\}" variable is replaced with the target movie plot, and "\{icl\_example\}" is filled with the list of in-context examples described with the template in \autoref{prompt:base-ice-template}.\par
For \rstos and \stos settings, the "Review" part of the output format specifier is removed. For \stosnospace, the "Review" part of the in-context example template is removed.
Note that newlines are inserted accordingly on the paper to improve the visibility. When applying the prompt to other datasets, we replace words representing the target dataset's domain. The tags like "<|start\_header\_id|>" are also replaced for experiments with different models.

\subsection{Prompt Engineering Techniques}
\label{appendix:prompt-engineering}
\begin{figure}
\begin{Prompt}
\begin{verbatim}
A critic's past movie reviews are listed
below:

{icl_example}

Based on this user’s past reviews, what
are the most common scores they give
for positive and negative reviews?
Answer in the following form:

most common positive score:
<most common positive score>,
most common negative score:
<most common negative score>
\end{verbatim}
\end{Prompt}
\caption{Prompt used to generate the score range summarization text}
\label{prompt:scoresumm-template}
\end{figure}

\begin{figure}
\begin{Prompt}
\begin{verbatim}
A critic's past movie reviews
are listed below:

{icl_example}

Analyze the critic's preferences.
Provide clear explanations based
on details from the past reviews
and other pertinent factors.
\end{verbatim}
\end{Prompt}
\caption{Prompt used to generate the preference summary}
\label{prompt:profile-generation-template}
\end{figure}

\begin{figure}
\begin{Prompt}
\begin{verbatim}
The description of a movie plot is as follows:

{plot}

what else should I say if I want to
recommend it to others?
\end{verbatim}
\end{Prompt}
\caption{Prompt used to generate the item recommendation text}
\label{prompt:item-recommendation-template}
\end{figure}

In this section, we introduce detailed prompt templates used for experiments in \ref{subsection:other-prompt-engineering}
\paragraph{Zero-shot CoT}
We reused the prompt in \autoref{prompt:base-query-prompt}, except that the beginning of the assistant response is replaced with ``Let's think step by step.''.
\paragraph{Score Range Summary}
We use the prompt presented in \autoref{prompt:scoresumm-template} adopted from \citet{intsum} to generate the score range summary text, then add this intermediate output to the prompt in \autoref{prompt:base-query-prompt} with the prefix ``The trend of review scores given by this user is analyzed as follows:''
\paragraph{Preference Summary}
We use the prompt presented in \autoref{prompt:user-profile-template}, originally used for KAR~\cite{kar}, to generate the analysis of the user preference. This output is added to the rating prediction prompt in \autoref{prompt:base-query-prompt} with the prefix ``The preference of him/her is analyzed as follows:''.
\paragraph{Preference Summary + Item Recommendation}
In addition to the Preference Summary, we also add the item recommendation text generated with the prompt presented in \autoref{prompt:item-recommendation-template}, which is originally used in LLM-Rec\cite{llm-rec}.\par
Then the item recommendation text is also added to the bottom of the prompt in \autoref{prompt:base-query-prompt}, surrounded by ``[The Start of Recommendation Text]'' and ``[The End of Recommendation Text]'' tags.

\subsection{Self-Described Preference}
\label{appendix-self-description}
\begin{figure}[t]
  \centering
    \includegraphics[width=1.0\columnwidth]{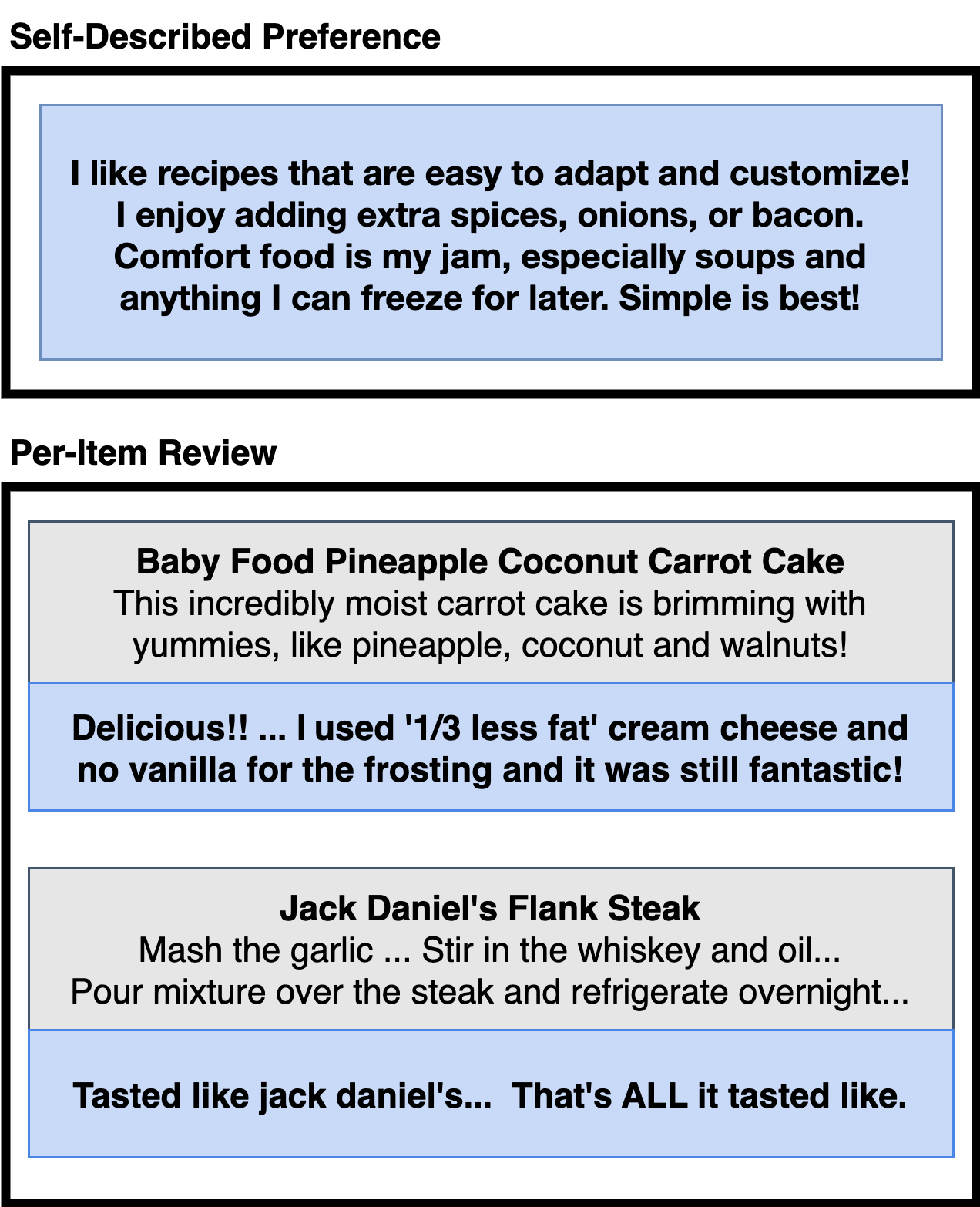}
  \caption{Comparison between self-described preference (top) and per-item review (bottom). Per-item review format can contain more specific preference information, and makes it easy to add more information if available.}
  \label{fig:self-description-diagram}
\end{figure}

\begin{table*}[tb]
  \centering
  \begin{tabularx}{\textwidth}{@{}lX@{}}
    \toprule
    Original Example  &
      I like comedy genre movies, while watching comedy movies I will feel very happy and relaxed. Comedy films are designed to make the audience laugh. It has different kinds of categories in comedy genres such as horror comedy, romantic comedy, comedy thriller, musical-comedy. \\
    \bottomrule
  \end{tabularx}
  \caption{Example in the original dataset proposed by \citet{profile-based-recommendation}}
  \label{tab:original-description}
\end{table*}

\begin{table*}[tb]
  \centering
  \begin{tabularx}{\textwidth}{@{}lX@{}}
    \toprule
    Gemma 3 12B   &
      I like complex plots with suspense, intrigue, and a touch of action. Gritty noir films and thrillers with morally ambiguous characters are right up my alley! A good story is key. \\
    \addlinespace
    Llama 3.1 8B  &
      I like complex, suspenseful stories with intricate plots and unexpected twists. I'm drawn to films that explore the human condition, morality, and the blurred lines between right and wrong. I appreciate gritty, atmospheric settings and powerful filmmaking. \\
    \bottomrule
  \end{tabularx}
  \caption{Examples of self-description style preference generated by LLMs}
  \label{tab:llm-description}
\end{table*}

\begin{figure}
\begin{Prompt}
\begin{verbatim}
A critic's past movie reviews are listed
below:

{icl_example}

Write the passage this person would write when
asked to describe their movie preferences.
The passage must start with “I like …” and
be no more than 300 characters long.
\end{verbatim}
\end{Prompt}
\caption{Prompt used to convert the per-item review to self-description style text}
\label{prompt:user-profile-template}
\end{figure}

\autoref{fig:self-description-diagram} illustrates the difference between per-item review and self-described preference data. 
For the experiments in \autoref{self-description}, we use the prompt in \autoref{prompt:profile-generation-template} to transform the per-item review text into the self-description style text. Example texts are listed in \autoref{tab:llm-description}. LLMs successfully generate the self-description style text similar to the original example of \citet{profile-based-recommendation} presented in \autoref{tab:original-description}.\par
At inference time, the self-description text is added to the review prediction prompt in \autoref{prompt:base-query-prompt} with the prefix "His / her self-description of the preference is as follows:".

\section{Detailed Results}
\subsection{Detailed Results of \autoref{subsection:base-prompting-results}}
\label{appendix:results-detail}
\begin{table*}[ht]
  \centering
  \resizebox{\textwidth}{!}{%
  \begin{tabular}{@{}llrrrrrrrr@{}}  %
    \toprule
    Dataset & Model 
    & \multicolumn{4}{c}{S $\to$ S}
    & \multicolumn{4}{c}{RS $\to$ S}\\
    \cmidrule(lr){3-6}\cmidrule(lr){7-10}
    &      & $\rho$ & $\tau$ & RMSE & FR    
           & $\rho$ & $\tau$ & RMSE & FR \\
    \midrule
Movies & LLama 3.1 8B
	& 0.140 & 0.117 & 2.722 & 0
	& \textbf{0.212} & \textbf{0.176} & \textbf{2.429} & 0
\\
 & LLama 3.3 70B
	& 0.265 & 0.213 & 2.408 & 0
	& \textbf{0.293} & \textbf{0.239} & \textbf{2.276} & 0
\\
 & Gemma 3 12B
	& 0.163 & 0.131 & 2.539 & 0
	& \textbf{0.239} & \textbf{0.190} & \textbf{2.383} & 0.001
\\
 & Gemma 3 27B
	& 0.206 & 0.163 & 2.503 & 0
	& \textbf{0.237} & \textbf{0.188} & \textbf{2.470} & 0.002
\\
 & Qwen 3 8B
	& 0.144 & 0.120 & 2.303 & 0
	& \textbf{0.198} & \textbf{0.161} & \textbf{2.259} & 0
\\
 & o3
	& 0.341 & 0.270 & 2.351 & 0
	& \textbf{0.488} & \textbf{0.395} & \textbf{2.046} & 0
\\
 & GPT-4.1
	& 0.340 & 0.270 & 2.307 & 0
	& \textbf{0.443} & \textbf{0.359} & \textbf{2.037} & 0
\\
 & Claude 4 Sonnet
	& 0.301 & 0.236 & 2.491 & 0.004
	& \textbf{0.388} & \textbf{0.306} & \textbf{2.371} & 0.028
\\
 & User Average
	& 0.231 & 0.172 & 2.241 & 0
	& 0.231 & 0.172 & 2.241 & 0
\\
 & Matrix Factorization
	& 0.557 & 0.427 & 2.086 & 0
	& 0.557 & 0.427 & 2.086 & 0
\\
\midrule
Recipe & LLama 3.1 8B
	& 0.064 & 0.063 & 0.726 & 0
	& \textbf{0.096} & \textbf{0.094} & \textbf{0.723} & 0.000
\\
 & LLama 3.3 70B
	& 0.152 & 0.148 & 0.736 & 0
	& \textbf{0.160} & \textbf{0.156} & \textbf{0.716} & 0
\\
 & Gemma 3 12B
	& 0.161 & 0.155 & 0.787 & 0
	& \textbf{0.205} & \textbf{0.198} & \textbf{0.755} & 0
\\
 & Gemma 3 27B
	& 0.161 & 0.155 & \textbf{0.782} & 0
	& \textbf{0.208} & \textbf{0.199} & 0.799 & 0
\\
 & Qwen 3 8B
	& 0.224 & 0.219 & \textbf{0.694} & 0
	& \textbf{0.241} & \textbf{0.234} & 0.709 & 0
\\
 & o3
	& 0.222 & 0.213 & 0.787 & 0
	& \textbf{0.244} & \textbf{0.237} & \textbf{0.727} & 0
\\
 & GPT-4.1
	& 0.182 & 0.174 & 0.844 & 0
	& \textbf{0.207} & \textbf{0.200} & \textbf{0.748} & 0
\\
 & Claude 4 Sonnet
	& 0.207 & 0.196 & 0.879 & 0
	& \textbf{0.234} & \textbf{0.226} & \textbf{0.758} & 0.023
\\
 & User Average
	& 0.270 & 0.236 & 0.651 & 0
	& 0.270 & 0.236 & 0.651 & 0
\\
 & Matrix Factorization
	& 0.061 & 0.048 & 1.512 & 0
	& 0.061 & 0.048 & 1.512 & 0
\\
\midrule
Books & LLama 3.1 8B
	& 0.171 & 0.159 & 1.374 & 0
	& \textbf{0.186} & \textbf{0.173} & \textbf{1.326} & 0
\\
 & LLama 3.3 70B
	& 0.255 & 0.235 & \textbf{1.395} & 0
	& \textbf{0.260} & \textbf{0.240} & 1.405 & 0
\\
 & Gemma 3 12B
	& 0.221 & 0.198 & 1.424 & 0
	& \textbf{0.308} & \textbf{0.276} & \textbf{1.265} & 0.000
\\
 & Gemma 3 27B
	& 0.226 & 0.204 & 1.405 & 0
	& \textbf{0.296} & \textbf{0.263} & \textbf{1.366} & 0.002
\\
 & Qwen 3 8B
	& 0.283 & 0.260 & 1.216 & 0
	& \textbf{0.298} & \textbf{0.272} & \textbf{1.175} & 0
\\
 & o3
	& 0.280 & 0.251 & 1.361 & 0
	& \textbf{0.331} & \textbf{0.301} & \textbf{1.201} & 0
\\
 & GPT-4.1
	& 0.267 & 0.240 & 1.417 & 0
	& \textbf{0.308} & \textbf{0.278} & \textbf{1.276} & 0
\\
 & Claude 4 Sonnet
	& 0.205 & 0.181 & 1.593 & 0.002
	& \textbf{0.315} & \textbf{0.280} & \textbf{1.340} & 0.003
\\
 & User Average
	& 0.353 & 0.292 & 1.126 & 0
	& 0.353 & 0.292 & 1.126 & 0
\\
 & Matrix Factorization
	& 0.234 & 0.181 & 1.812 & 0
	& 0.234 & 0.181 & 1.812 & 0
\\

\bottomrule
  \end{tabular}%
  }
\caption{Comparison of \stos and \rstors prompting. Symbols: $\rho$ = Spearman correlation, $\tau$ = Kendall--$\tau$ correlation, FR = failure rate.}
  \label{tab:base-results}
\end{table*}

We report the concrete numbers of Spearman Correlation, Kendall-Tau correlation, RMSE, and Failure Rate of the experiment of \autoref{subsection:base-prompting-results} in \autoref{tab:base-results}. The failure rate is highest (2.8\%) with the combination of Claude 4 Sonnet and the Movies dataset, but generally at an acceptable level.\par
A reduction in RMSE is not observed for three model-dataset combinations (Gemma 3 27B and Qwen3 8B on Recipe; Llama 3.3 70B on Books). Furthermore, the improvement in Spearman correlation for one model (Llama 3.1 8B) is not statistically significant.

A potential explanation for these exceptions is the skewed ground-truth label distribution in the Recipe and Books datasets, as shown in Appendix \ref{appendix:dataset-stats}.
For such datasets, a naive heuristic of consistently predicting a high score can outperform the content-based reasoning.
Nevertheless, it is important to note that no substantial performance degradation is observed in these cases.
The negative effects are negligible, especially when compared with the overall benefits of using in-context reviews.

\subsection{Detailed Results of \autoref{subsection:prompt-engineering-result}}
\label{appendix:prompt-engineering-results-detail}
\begin{table*}[ht]
  \centering
  \resizebox{\textwidth}{!}{%
  \begin{tabular}{@{}llrrrrrrrr@{}}  %
    \toprule
    Dataset & Model 
    & \multicolumn{4}{c}{RS $\to$ S}
    & \multicolumn{4}{c}{RS $\to$ RS}\\
    \cmidrule(lr){3-6}\cmidrule(lr){7-10}
    &      & $\rho$ & $\tau$ & RMSE & FR    
           & $\rho$ & $\tau$ & RMSE & FR \\
    \midrule
Movies & LLama 3.1 8B
	& 0.212 & 0.176 & \textbf{2.429} & 0
	& \textbf{0.241} & \textbf{0.188} & 2.609 & 0.005
\\
 & LLama 3.3 70B
	& 0.293 & 0.239 & \textbf{2.276} & 0
	& \textbf{0.305} & \textbf{0.250} & 2.451 & 0
\\
 & Gemma 3 12B
	& 0.239 & 0.190 & \textbf{2.383} & 0.001
	& \textbf{0.274} & \textbf{0.217} & 2.472 & 0.002
\\
 & Gemma 3 27B
	& 0.237 & 0.188 & 2.470 & 0.002
	& \textbf{0.251} & \textbf{0.199} & \textbf{2.460} & 0.002
\\
 & Qwen 3 8B
	& 0.198 & 0.161 & 2.259 & 0
	& \textbf{0.245} & \textbf{0.199} & \textbf{2.256} & 0
\\
 & o3
	& \textbf{0.488} & \textbf{0.395} & \textbf{2.046} & 0
	& 0.485 & 0.394 & 2.072 & 0.013
\\
 & GPT-4.1
	& \textbf{0.443} & \textbf{0.359} & \textbf{2.037} & 0
	& 0.390 & 0.313 & 2.234 & 0
\\
 & Claude 4 Sonnet
	& 0.388 & 0.306 & 2.371 & 0.028
	& \textbf{0.421} & \textbf{0.334} & \textbf{2.273} & 0.003
\\
\midrule
Recipe & LLama 3.1 8B
	& 0.096 & 0.094 & \textbf{0.723} & 0.000
	& \textbf{0.171} & \textbf{0.166} & 0.724 & 0.008
\\
 & LLama 3.3 70B
	& 0.160 & 0.156 & 0.716 & 0
	& \textbf{0.164} & \textbf{0.160} & \textbf{0.715} & 0.002
\\
 & Gemma 3 12B
	& 0.205 & 0.198 & 0.755 & 0
	& \textbf{0.223} & \textbf{0.217} & \textbf{0.734} & 0.000
\\
 & Gemma 3 27B
	& \textbf{0.208} & \textbf{0.199} & 0.799 & 0
	& 0.203 & 0.196 & \textbf{0.755} & 0.005
\\
 & Qwen 3 8B
	& \textbf{0.241} & \textbf{0.234} & 0.709 & 0
	& 0.181 & 0.176 & \textbf{0.698} & 0.001
\\
 & o3
	& \textbf{0.244} & \textbf{0.237} & 0.727 & 0
	& 0.215 & 0.210 & \textbf{0.700} & 0
\\
 & GPT-4.1
	& \textbf{0.207} & \textbf{0.200} & 0.748 & 0
	& 0.196 & 0.191 & \textbf{0.707} & 0.003
\\
 & Claude 4 Sonnet
	& \textbf{0.234} & \textbf{0.226} & 0.758 & 0.023
	& 0.225 & 0.219 & \textbf{0.723} & 0.020
\\
\midrule
Books & LLama 3.1 8B
	& 0.186 & 0.173 & 1.326 & 0
	& \textbf{0.276} & \textbf{0.249} & \textbf{1.314} & 0.014
\\
 & LLama 3.3 70B
	& \textbf{0.260} & \textbf{0.240} & 1.405 & 0
	& 0.259 & 0.238 & \textbf{1.300} & 0.006
\\
 & Gemma 3 12B
	& 0.308 & 0.276 & 1.265 & 0.000
	& \textbf{0.330} & \textbf{0.295} & \textbf{1.257} & 0.003
\\
 & Gemma 3 27B
	& 0.296 & 0.263 & 1.366 & 0.002
	& \textbf{0.302} & \textbf{0.270} & \textbf{1.301} & 0.008
\\
 & Qwen 3 8B
	& \textbf{0.298} & \textbf{0.272} & \textbf{1.175} & 0
	& 0.271 & 0.252 & 1.215 & 0
\\
 & o3
	& 0.331 & 0.301 & 1.201 & 0
	& \textbf{0.368} & \textbf{0.339} & \textbf{1.149} & 0.001
\\
 & GPT-4.1
	& 0.308 & 0.278 & 1.276 & 0
	& \textbf{0.342} & \textbf{0.312} & \textbf{1.225} & 0
\\
 & Claude 4 Sonnet
	& 0.315 & 0.280 & 1.340 & 0.003
	& \textbf{0.316} & \textbf{0.283} & \textbf{1.302} & 0.004
\\

\bottomrule
  \end{tabular}%
  }
\caption{Comparison of \rstos and \rstors prompting. Symbols: $\rho$ = Spearman correlation, $\tau$ = Kendall--$\tau$ correlation, FR = failure rate.}
  \label{tab:prompt-engineering-results}
\end{table*}

 \autoref{tab:prompt-engineering-results} shows the detailed numeric results of \autoref{subsection:base-prompting-results}.

\subsection{Output Distribution of different models}
\label{appendix:output-distribution}
\begin{figure}[ht]
  \centering
  \begin{subfigure}[b]{0.4\linewidth}
    \includegraphics[width=\linewidth]{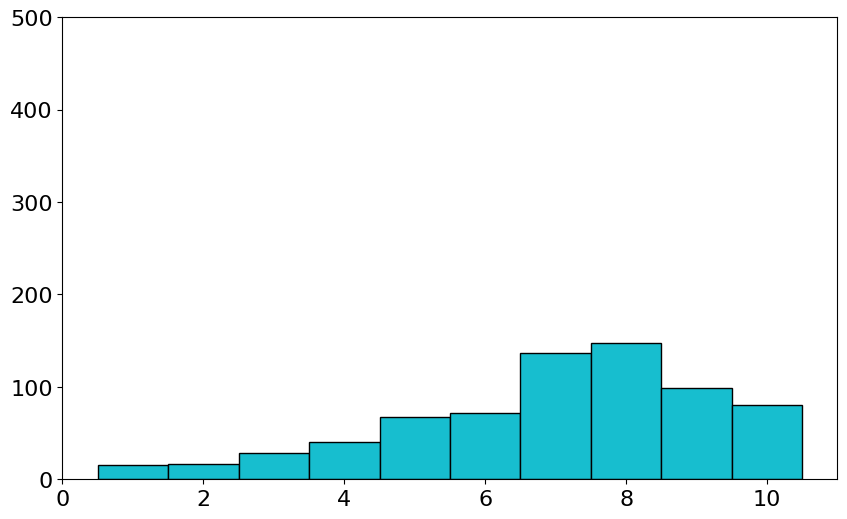}
    \caption{Ground Truth}
    \label{fig:llama-ground-truth-distr}
  \end{subfigure}%
  \quad
  \begin{subfigure}[b]{0.4\linewidth}
    \includegraphics[width=\linewidth]{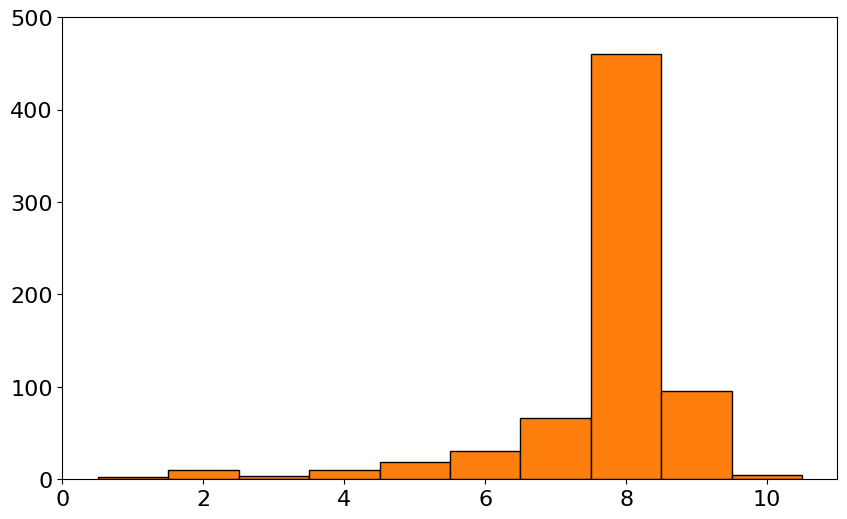}
    \caption{\rstos}
    \label{fig:llama-rss-distr}
  \end{subfigure}

  \vspace{0.3em}

  \begin{subfigure}[b]{0.4\linewidth}
    \includegraphics[width=\linewidth]{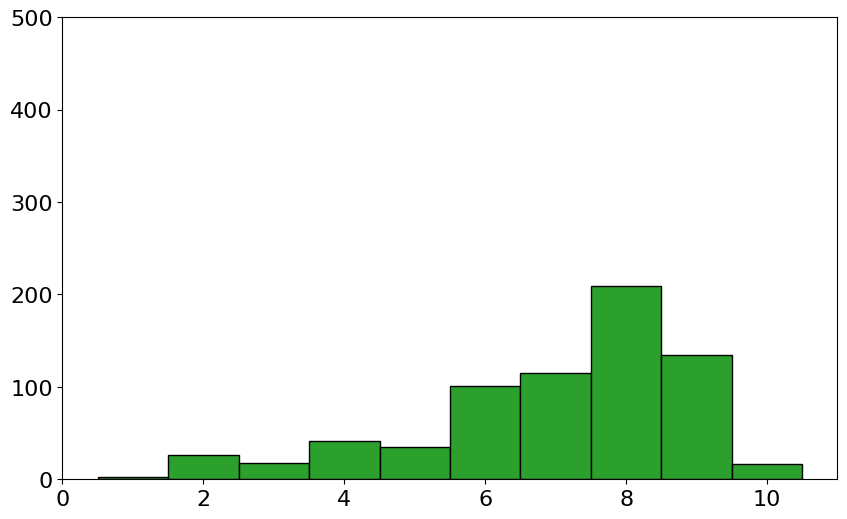}
    \caption{\rstors}
    \label{fig:llama-rsrs-distr}
  \end{subfigure}%
  \quad
  \begin{subfigure}[b]{0.4\linewidth}
    \includegraphics[width=\linewidth]{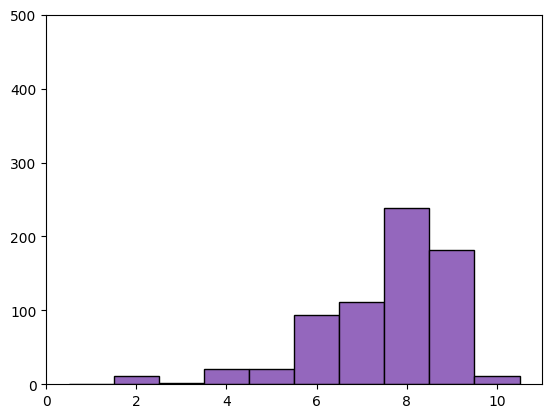}
    \caption{\rstors +  CoT}
    \label{fig:llama-cot-distr}
  \end{subfigure}

  \caption{Output rating distribution of Llama 3.1 8B on the Movies dataset with different prompting methods. The distribution change is similar to that of Gemma 3 12B.}
  \label{fig:llama-prompt-engineering-label-stats}
\end{figure}

\begin{figure}[ht]
  \centering
  \begin{subfigure}[b]{0.4\linewidth}
    \includegraphics[width=\linewidth]{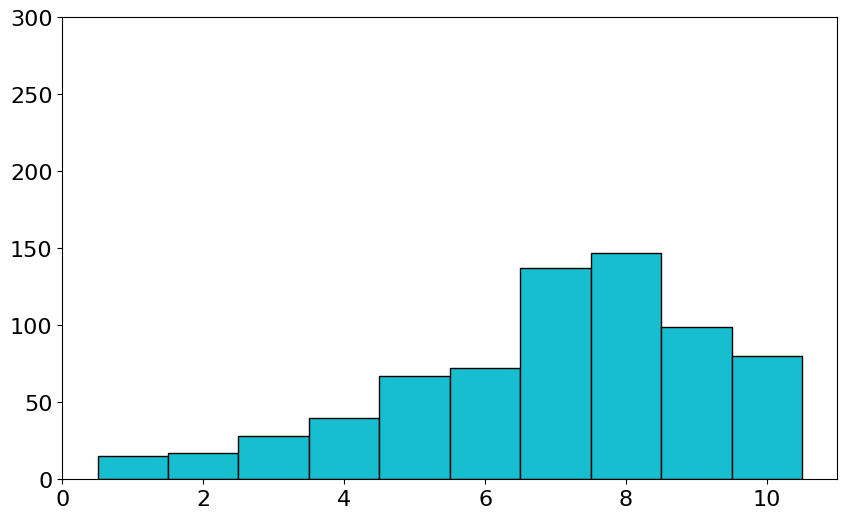}
    \caption{Ground Truth}
    \label{fig:o3-ground-truth-distr}
  \end{subfigure}%
  \quad
  \begin{subfigure}[b]{0.4\linewidth}
    \includegraphics[width=\linewidth]{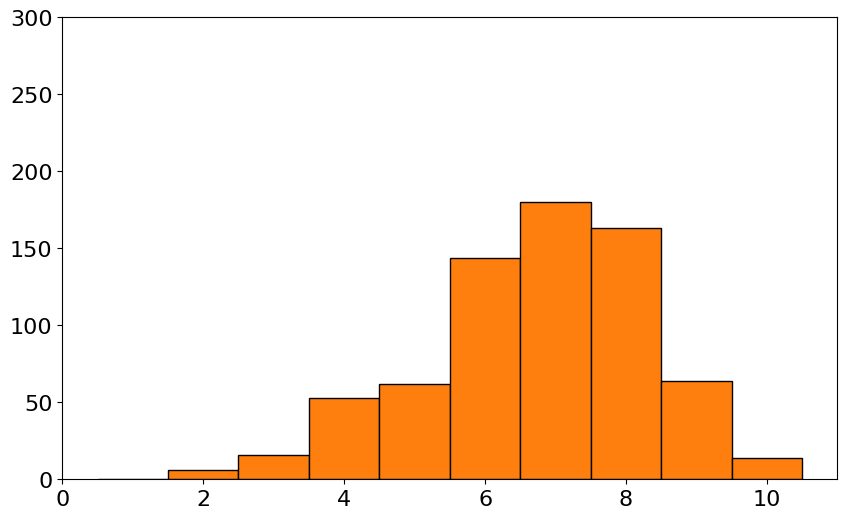}
    \caption{\rstos}
    \label{fig:o3-rss-distr}
  \end{subfigure}

  \vspace{0.3em}

  \begin{subfigure}[b]{0.4\linewidth}
    \includegraphics[width=\linewidth]{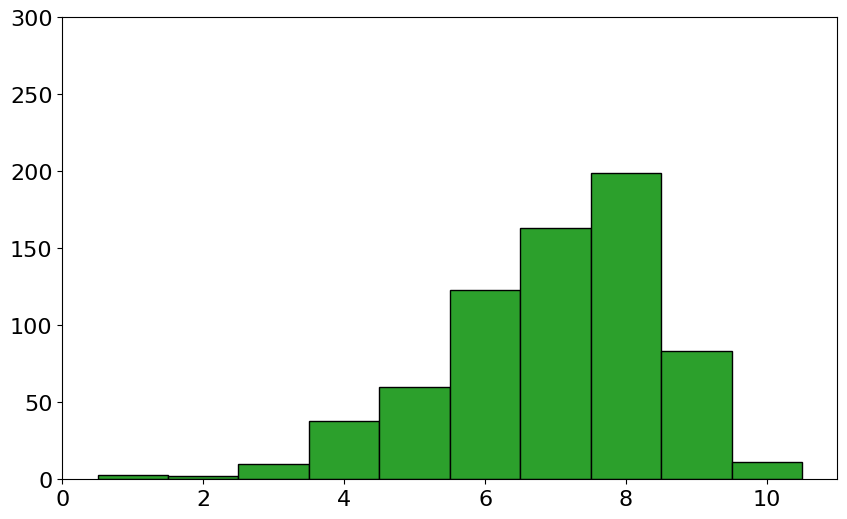}
    \caption{\rstors}
    \label{fig:o3-rsrs-distr}
  \end{subfigure}%
  \caption{Output rating distribution of o3 on the Movies dataset with different prompting methods. There is not a huge output distrbution shift as the prompting method changes.}
  \label{fig:o3-prompt-engineering-label-stats}
\end{figure}

\autoref{fig:llama-prompt-engineering-label-stats} and \autoref{fig:o3-prompt-engineering-label-stats} show the output distribution of Llama 3.1 8B and o3 on the Movies dataset with different prompting methods respectively. While Llama 3.1 8B shows a similar distribution shift as that of Gemma 3 12B, o3 does not show a significant change as the prompting method changes.

\subsection{Concrete Outputs with Different Prompting Styles}
\label{appendix-prompt-engineering-results}
\begin{table*}[tb]
  \centering
  \begin{tabularx}{\textwidth}{@{}llX@{}}
    \toprule
    Prompting Method & Score & Raw Response \\
    \hline
    Ground Truth & 1 / 10   &
      this crap gives Dracula a bad name ||| This is one of the most inane films I have ever had the misfortune of viewing... \\
    \addlinespace
    \rstos & 7 /10  &
      N/A \\
        \addlinespace
    \rstors & 3 /10  &
      ""Review"": ""What a load of hooey! ||| This one was a real mess. Too many characters, too many ridiculous plot twists... \\
    \addlinespace
    \rstors + Zero-shot CoT & 8 /10  &
      Okay, analyzing the critic's preferences:... \newline *   **Dislikes:** ""Sugary,"" overly sentimental/romantic ... \newline *   **Likes:** Strong characters...
      \newline ""Review"": ""Another bloodsucker on the loose ||| Well, at least this one doesn't insult the viewer's intelligence too much...\\
    \bottomrule
  \end{tabularx}
  \caption{Example responses by Gemma 3 12B on the Movies dataset with different prompting methods}
  \label{tab:llm-responses}
\end{table*}

\autoref{tab:llm-responses} lists the outputs on a data point in the Movies dataset by Gemma 3 12B, based on different prompting styles. As the table shows, with \rstos the model predicts seven as a generally plausible score, while with \rstors the model predicts three, which is close to the ground truth score. However, when zero-shot CoT is also applied, the model lists up the user's dislikes and likes first, and predicts a more favorable score of eight. This example aligns with the output distribution change illustrated in \autoref{fig:gemma-prompt-engineering-label-stats}.

\section{License and Intended Use of Scientific Artifacts}
In this work, scientific artifacts including datasets (\autoref{base-datasets}), models (\autoref{models}), and other software (Appendix \ref{appendix-other-software}) are used under the specified license and the terms of use. 

\section{AI Assistance Usage}
In this work, ChatGPT\footnote{https://chatgpt.com/} has been used for writing elaboration. GitHub Copilot\footnote{https://github.com/features/copilot} has also been used as a coding assistant for the experiments.

Google Gemini\footnote{https://gemini.google.com/} has been used for both purposes as well.

\end{document}